\documentclass[10pt,twocolumn,letterpaper]{article}

\usepackage[pagenumbers]{cvpr}              %
\usepackage{graphicx}
\usepackage{amsmath}
\usepackage{amssymb}
\usepackage{booktabs}
\usepackage{algorithm}
\usepackage{dsfont}
\usepackage{algorithm}
\usepackage{algpseudocode}
\usepackage{algorithmicx}
\usepackage{mathtools}
\usepackage{multirow}
\usepackage{graphics}
\usepackage{wrapfig}
\usepackage{bbm}
\usepackage{xspace}
\usepackage{xcolor}
\usepackage{tikz}

\newcommand{\methodName}{CorrespondentDream\xspace}

\usepackage{xr}
\makeatletter
\newcommand*{\addFileDependency}[1]{%
  \typeout{(#1)}
  \@addtofilelist{#1}
  \IfFileExists{#1}{}{\typeout{No file #1.}}
}
\makeatother

\definecolor{cvprblue}{rgb}{0.21,0.49,0.74}
\usepackage[pagebackref,breaklinks,colorlinks,citecolor=cvprblue]{hyperref}

\def\paperID{13247} %
\def\confName{CVPR}
\def\confYear{2024}

\title{Enhancing 3D Fidelity of Text-to-3D using Cross-View Correspondences}

\author{Seungwook Kim$^{1,2}$ \hspace{0.6cm} Kejie Li$^2$ \hspace{0.6cm} Yichun Shi$^2$ \hspace{0.6cm} Xueqing Deng$^2$ \hspace{0.6cm} Minsu Cho$^1$ \hspace{0.6cm} Peng Wang$^2$ \vspace{0.3cm}\\
$^1$POSTECH, South Korea \hspace{3.0cm} $^2$ByteDance, USA
}

\begin{document}

\twocolumn[{
\renewcommand\twocolumn[1][]{#1}%
\maketitle
\vspace{-8.0mm}
\includegraphics[width=.5\textwidth]{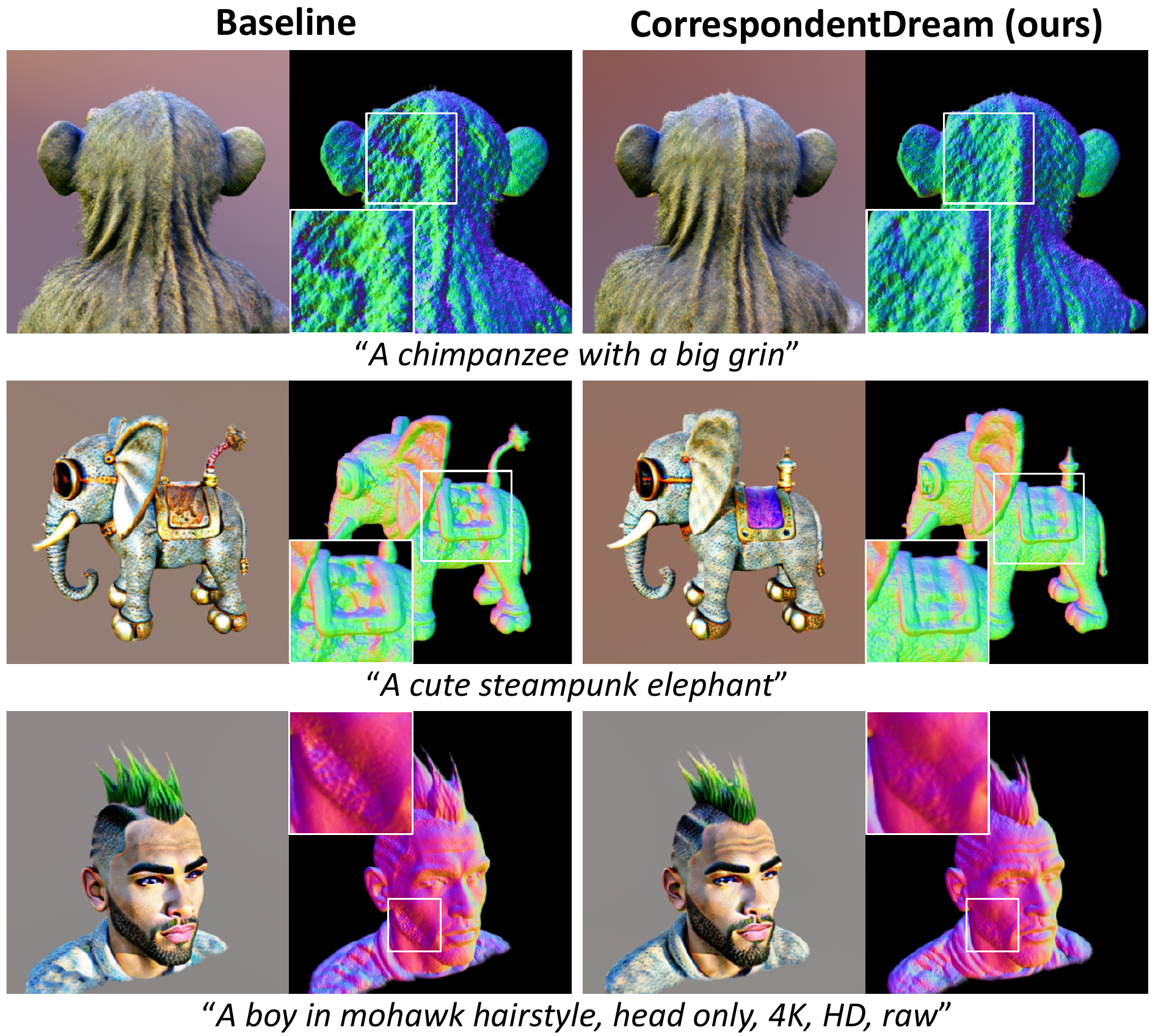}
\hfill
\includegraphics[width=.5\textwidth]{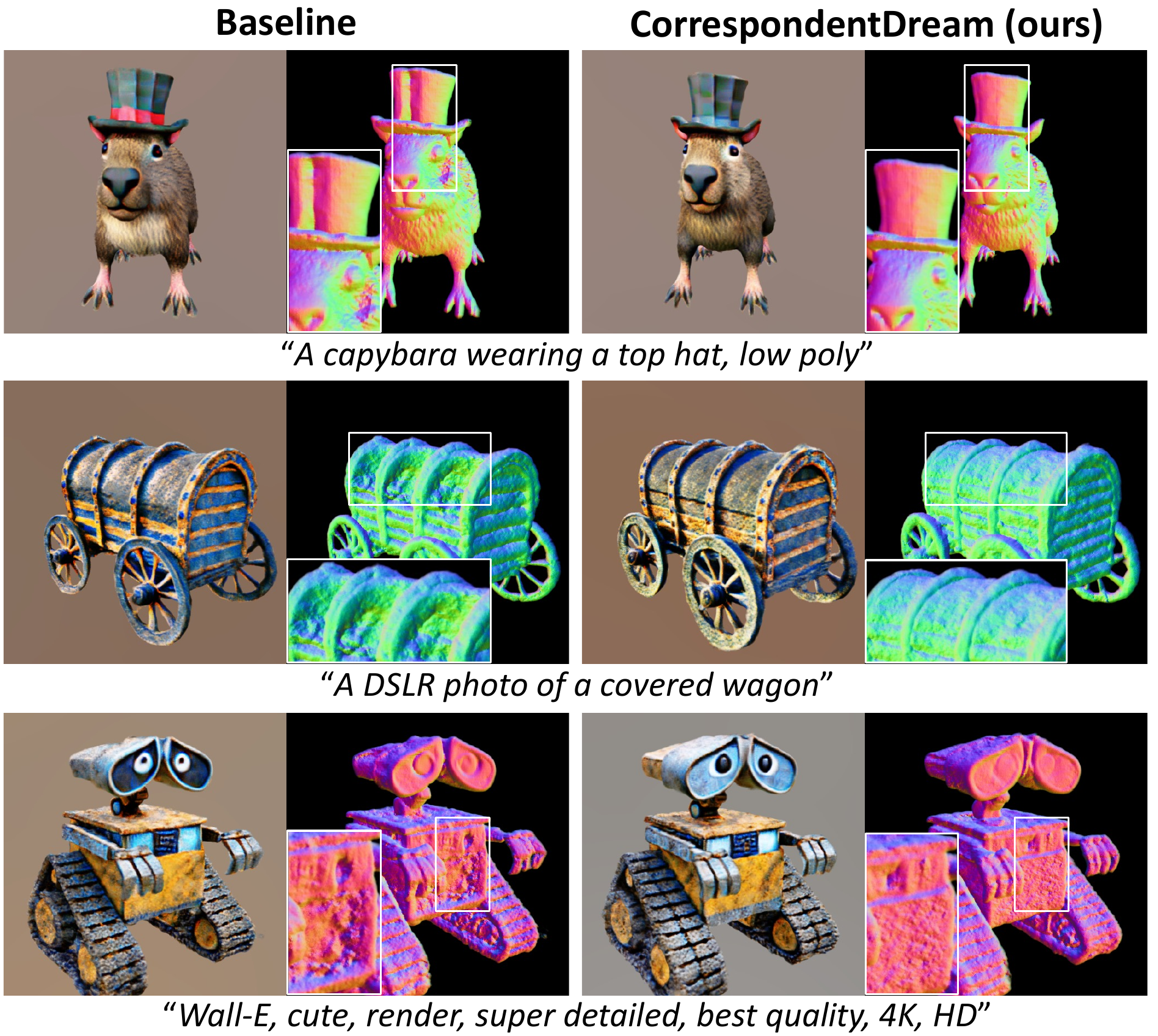}
\vspace{-5.5mm}
\captionof{figure}{\textbf{Comparison between the baseline (MVDream~\cite{shi2023mvdream}) and \methodName (ours).}
Our method substantially alleviates the 3D geometric infidelity issue in zero-shot text-to-3D generation methods.
Best viewed on electronics, zoom in for clearer visualization.
} 
\vspace{+4.0mm}
\label{fig:teaser}
}]

\begin{abstract}
Leveraging multi-view diffusion models as priors for 3D optimization have alleviated the problem of 3D consistency, \eg, the Janus face problem or the content drift problem, in zero-shot text-to-3D models.
However, the 3D geometric fidelity of the output remains an unresolved issue; albeit the rendered 2D views are realistic, the underlying geometry may contain errors such as unreasonable concavities.
In this work, we propose \methodName, an effective method to leverage annotation-free, cross-view correspondences yielded from the diffusion U-Net to provide additional 3D prior to the NeRF optimization process.
We find that these correspondences are strongly consistent with human perception, and by adopting it in our loss design, we are able to produce NeRF models with geometries that are more coherent with common sense, \eg, more smoothed object surface, yielding higher 3D fidelity.
We demonstrate the efficacy of our approach through various comparative qualitative results and a solid user study.
\end{abstract}
    
\section{Introduction}
\label{sec:intro}
Text-to-3D generation holds wide applicability in areas such as virtual reality and 3D content generation~\cite{poole2022dreamfusion}, which are of integral importance in the fields of gaming and media.
In recent studies, leveraging 2D diffusion models as priors to optimize 3D representations, \eg, NeRF~\cite{mildenhall2021nerf} or NeuS~\cite{wang2021neus}, via Score Distillation Sampling (SDS) has shown to yield promising results and generalizability for zero-shot text-to-3D generation~\cite{poole2022dreamfusion, lin2023magic3d}.

\begin{figure}[t]
    \begin{center}
        \includegraphics[width=\linewidth]{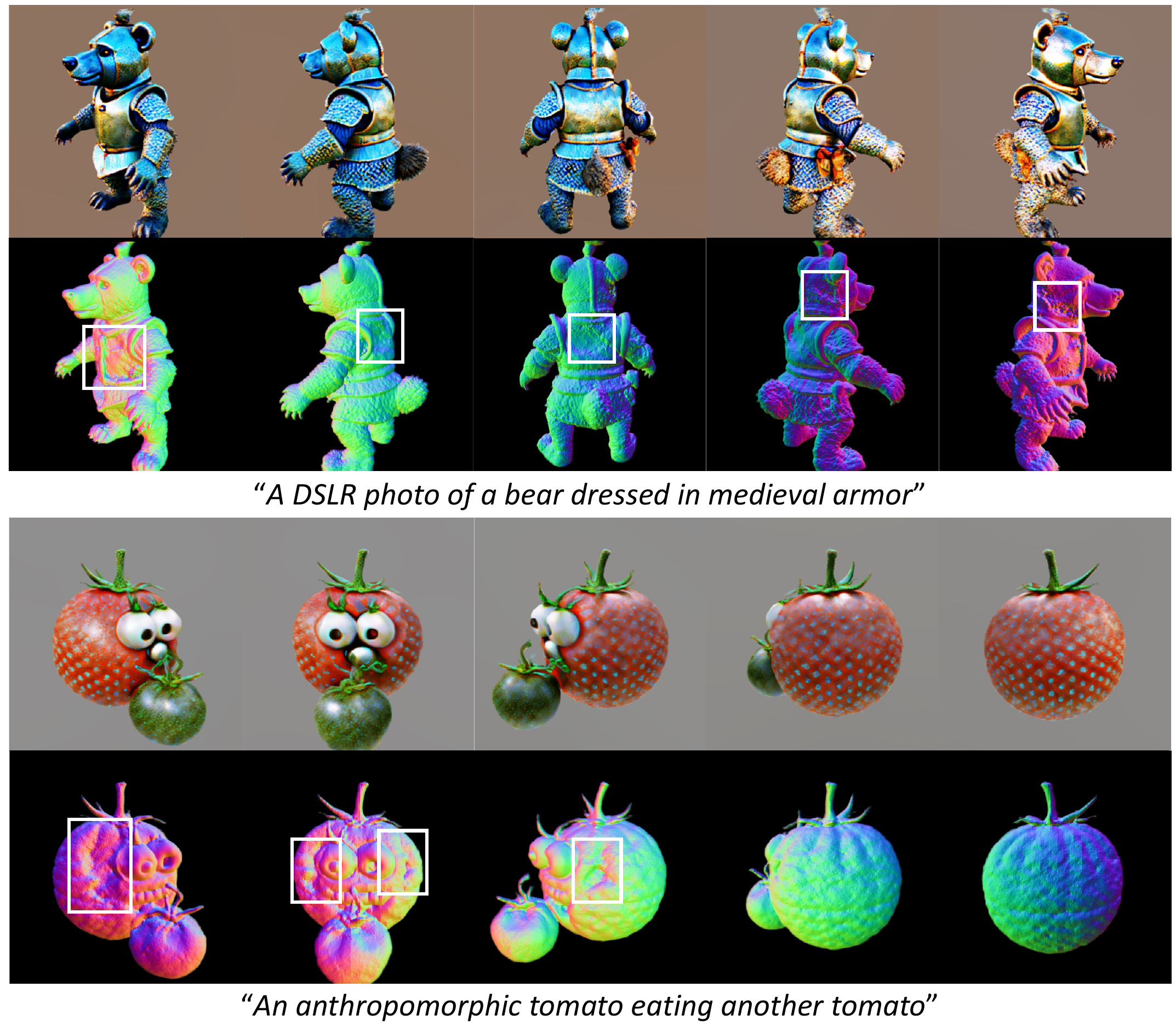}
    \end{center}
    \vspace{-5.0mm}
      \caption{\textbf{Rendered 2D view and 3D normal map of MVDream~\cite{shi2023mvdream}}. 
      While the rendered 2D views look realistic, the underlying 3D geometry lacks fidelity, with concavities or missing surfaces (highlighted in white squares).
      \vspace{-5.0mm}
}
\label{fig:3d_infidelity}
\end{figure}

It was subsequently observed that using a single-view 2D diffusion model as prior suffers from the lack of multi-view knowledge and 3D awareness, frequently resulting in issues including the Janus face problem or content drift~\cite{hong2023debiasing, shi2023mvdream}.
This problem was largely alleviated by leveraging a \textit{multi-view} diffusion model~\cite{shi2023mvdream} as the prior instead, which generates multiple view-consistent 2D images instead of a single 2D image to improve the multi-view consistency of the 3D output.
However, even with the integration of multi-view diffusion models, the disparity in dimensionality between the 2D priors and the final 3D representation makes it insufficient to ensure the 3D geometric fidelity of the output shape, as exemplified in~\cref{fig:3d_infidelity}.

In this paper, we introduce \methodName, a novel method which enhances 3D fidelity in text-to-3D generation, using \emph{cross-view correspondences} computed from the diffusion model functioning as the optimization prior.
By utilizing features from upsampling layers of the diffusion U-Net, we can establish robust correspondences between multi-view images without explicit supervision or fine-tuning.
Our approach hinges on the multi-view consistency of 2D features in the multi-view diffusion model, which we conjecture to be faithful to human perception.

By using the known camera parameters for NeRF-rendered views, we can reproject pixels across different views using the NeRF-rendered depth values.
In the presence of 3D infidelities such as concavities or missing surfaces, the NeRF-rendered depth values will also be erroneous, reflecting the infidelities. 
We aim to correct these errors by aligning the NeRF reprojections with cross-view correspondences, thereby enhancing the 3D fidelity of the output by correcting the NeRF depths.
The effectiveness of \methodName is validated through extensive qualitative assessments and a user study.

The contributions of our work are threefold:
\begin{itemize}
\item We identify that 3D infidelities remains an issue in existing zero-shot text-to-3D methods, even with improved 3D consistency via multi-view diffusion priors.
\item We introduce \methodName, a novel method to incorporate \textit{cross-view correspondences} into the 3D optimization for improved 3D fidelity.
\item We demonstrate the effectiveness of our method via comparative analysis and a user study.
\end{itemize}

\section{Related Work}
\label{sec:relatedwork}

\smallbreak
\noindent
\textbf{Text-to-3D using 2D diffusion models.}
Based on the observation that template-based generation pipelines and 3D generative models~\cite{chan2023genvs,henderson2020leveraging, wang2023rodin, karnewar2023holodiffusion} show limited 3D generation performances due to the lack of sufficiently large-scale 3D data, 2D-lifting methods have gained interest~\cite{poole2022dreamfusion,lin2023magic3d}.
Specifically, DreamFusion~\cite{poole2022dreamfusion} proposed Score Distillation Sampling (SDS) to leverage 2D diffusion models as priors to optimize 3D representations~\cite{mildenhall2021nerf,wang2021neus} to facilitate zero-shot text-to-3D generation, while SJC~\cite{wang2023score} concurrently proposed a similar technique using the stable-diffusion model~\cite{rombach2022high}.
Subsequent studies aim to improve the output representation~\cite{chen2023fantasia3d, tang2023make}, sampling schedules for optimization~\cite{huang2023dreamtime}, and loss design~\cite{wang2023prolificdreamer} for improved quality and efficiency.
However, using a single-view 2D diffusion prior is observed to suffer from multi-view inconsistency - namely the Janus face problem and content drift.

\smallbreak
\noindent
\textbf{Text-to-3D using multi-view diffusion models.} 
To alleviate the problem of multi-view inconsistency, a promising direction is to leverage improved multi-view knowledge.
To this end, MVDream~\cite{shi2023mvdream} finetunes the stable diffusion~\cite{rombach2022high} model to generate multi-view images instead of a single-view image.
This is facilitated by replacing the self-attention of the diffusion U-Net with multi-view attention, such that the multiple views can attend to one another for multi-view knowledge.
Using multi-view diffusion as the prior to facilitate text-to-3D generation shows highly improved multi-view consistency in the rendered views.

Albeit their efficacy in addressing multi-view inconsistency between 2D rendered views, multi-view diffusion models still fall short in fully capturing the true fidelity of the underlying 3D geometry.
In this work, we address this issue via integrating cross-view correspondences from the diffusion network to enforce additional geometric priors.

\smallbreak
\noindent
\textbf{Establishing correspondences using diffusion models.}
With recent advancements in diffusion models~\cite{ho2020denoising, nichol2021improved, rombach2022high}, the potential of their representational abilities triggered many applications to visual correspondence.
Unsupervised methods already exhibit competitive performances, relying on iterative refinement of features~\cite{hedlin2023unsupervised} or an additional feature extractor~\cite{zhang2023tale} for improved performance.
The performance gains were notably higher in a strongly-supervised training scheme, either via aggregating the multi-scale features from multiple timesteps~\cite{luo2023dhf}, or by optimizing pair-specific prompts for better matchable features~\cite{li2023sd4match}.

In this work, we take inspiration from DIFT~\cite{tang2023dift} to yield cross-view correspondences using diffusion features.

\smallbreak
\noindent
\textbf{Leveraging correspondences for NeRF optimization.}
NeRF~\cite{mildenhall2021nerf} encodes 3D scenes with a MLP, which can eventually be used for 2D view rendering.
In recent studies, photometric loss alone proved to be insufficient to train a NeRF model under challenging constraints, \eg, sparse input views~\cite{jain2021putting, deng2022depth, truong2023sparf} or erroneous camera poses~\cite{jeong2021self, truong2023sparf}.
Using off-the-shelf image matching models as additional priors has shown promising results under sparse viewpoints or noisy poses~\cite{jeong2021self, truong2023sparf}.
SCNeRF~\cite{jeong2021self} aims to minimize the projected ray distance of off-the-shelf correspondences, while SPARF~\cite{truong2023sparf} proposes to minimize the reprojection error between the off-the-shelf matches and the reprojected matches obtained using NeRF depths.

In contrast, we compute correspondences between NeRF-rendered views, instead of ground truth images.
Furthermore, we do not rely on off-the-shelf matching methods, but compute annotation-free cross-view correspondences from diffusion features to provide additional geometric priors in optimizing NeRF for improved 3D fidelity.

\section{Preliminary: Text-to-3D using Diffusion}
\label{preliminary}

DreamFusion~\cite{poole2022dreamfusion} introduced Score Distillation Sampling (SDS), which facilitates the optimization of differentiable image parameterizations (DIP) by using diffusion models to compute gradients in the form of:
\begin{equation}
\nabla_{\theta} \mathcal{L}_{\textrm{SDS}}(\phi, x = g(\theta)) \triangleq \mathbb{E}_{t, \epsilon} \left[ w(t) \left( \epsilon_{\phi}(z_t; y, t) - \nabla_{\theta} x \right) \right].
\label{eqn:singleview_sds}
\end{equation}
In this formulation, \( \theta \) denotes the parameters of the DIP, \( \phi \) represents the parameters of the diffusion model, and \( x \) signifies the image rendered by the DIP through the function \( g \).
The term \( w(t) \) is a weighting function dependent on the sampled timestep \( t \).
The variable \( \epsilon \) stands for the noise vector, and \( z_t \) is the noisy image at timestep \( t \).
The expectation \( \mathbb{E} \) is taken over both \( t \) and \( \epsilon \), with \( y \) being the conditioning variable, such as a text prompt.
This approach allows a 2D diffusion model to act as a 'frozen critic', predicting image-space modifications to optimize the DIP. 
Common DIPs include 3D volumetric representations such as NeRF~\cite{mildenhall2021nerf} or NeuS~\cite{wang2021neus}, thus enabling zero-shot text-to-3D generation. 

However, a challenge arises from the lack of 3D consistency across rendered views due to the absence of integrated multi-view knowledge.
Research has shown promising results in addressing this challenge by introducing multi-view attention within the diffusion U-Nets to facilitate the training of multi-view diffusion models.
These models yield consistent multi-view color images, substantially improving 3D coherency~\cite{szymanowicz2023viewset, liu2023syncdreamer, shi2023mvdream}.
With a multi-view diffusion model at disposal, MVDream~\cite{shi2023mvdream} defines the multi-view diffusion loss for supervising the 3D volume as:
\begin{equation}
\mathcal{L}_{\textrm{SDS}}(\phi, \{x_i = g(\phi, c_i)\}_{i=1}^{N}) = \mathbb{E}_{t,c,\epsilon} \left[ \sum_{i=1}^{N} \| x_i - \hat{x}_{0,i} \|^2 \right]
\label{eqn:multiview_sds}
\end{equation}
Here, \( c_i \) denotes the camera pose for the \( i \)-th view, \( x_i \) is the image rendered from the 3D volume for the \( i \)-th view, and \( \hat{x}_{0,i} \) is the corresponding image generated by the diffusion model.
This deviates from the original SDS formulation (\cref{eqn:singleview_sds}); MVDream proposes that ~\cref{eqn:multiview_sds} is equivalent to ~\cref{eqn:singleview_sds} with $w(t)$ and shows to perform similarly as well, but using ~\cref{eqn:multiview_sds} further enables the usage of CFG rescale trick~\cite{lin2023common} to mitigate color saturation.
The improved 3D consistency afforded by these multi-view diffusion models ensures that the 3D volume's rendered views adhere to the same level of coherence, leading to substantial improvements in the stability and quality of text-to-3D generation.

\begin{figure*}[ht]
    \begin{center}
        \includegraphics[width=1.0\linewidth]{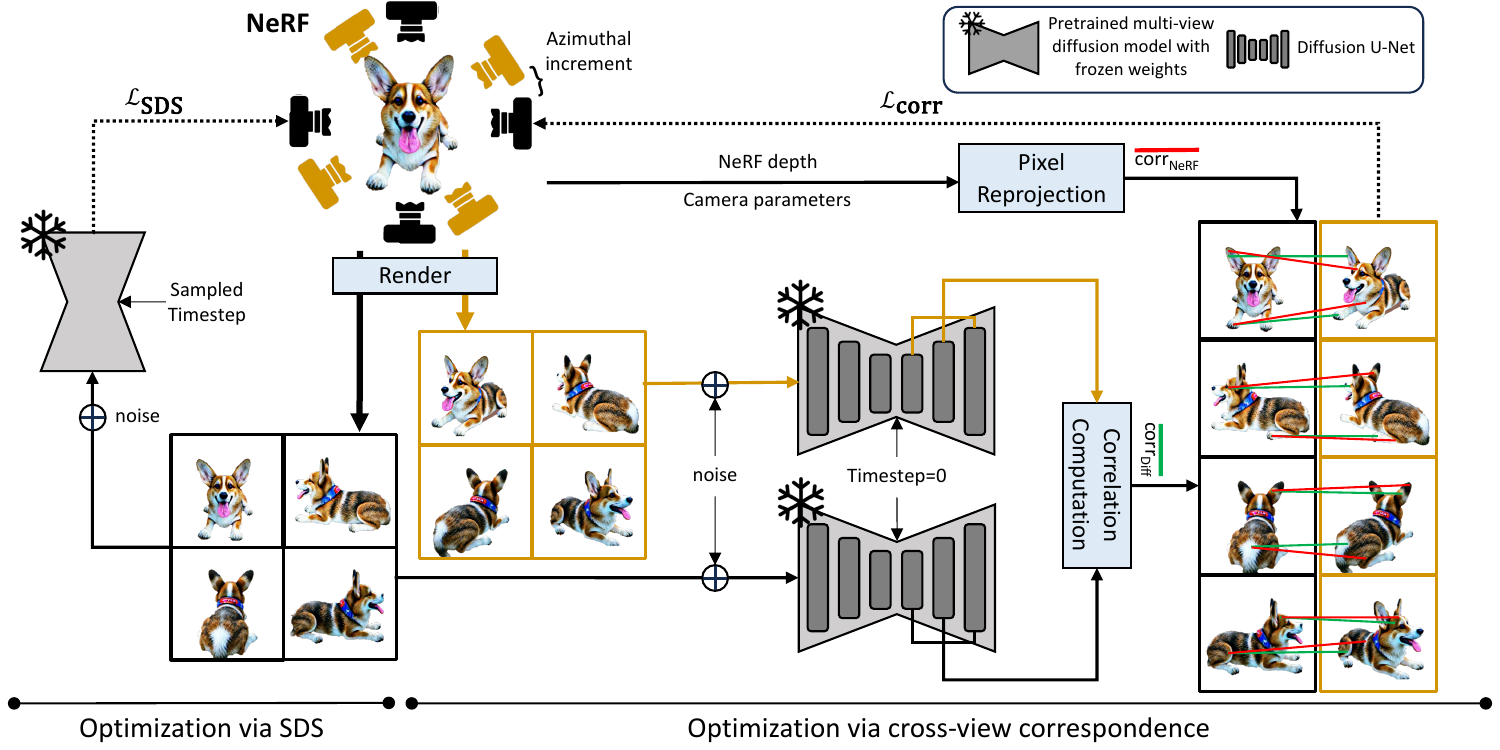}
    \end{center}
    \vspace{-5.0mm}
      \caption{\textbf{Overview of \methodName}. 
      We employ NeRF~\cite{mildenhall2021nerf} for 3D representation, optimized alternately using the SDS loss ($\mathcal{L}_{\textrm{SDS}}$) and cross-view correspondence loss ($\mathcal{L}_{\text{corr}}$).
      The $\mathcal{L}_{\textrm{SDS}}$ is based on the multi-view formulation from~\cref{eqn:multiview_sds} in MVDream~\cite{shi2023mvdream}.
      To compute $\mathcal{L}_{\text{corr}}$, we render two adjacent view sets from NeRF with identical noise, inputting them into a frozen pre-trained multi-view diffusion model.
      We then extract multi-layer features from the diffusion U-Net's upsampling layers to establish correspondences ($\text{corr}_\text{diff}$) between each view pair.
      Utilizing ground-truth camera parameters and NeRF-rendered depth, we reproject pixels to obtain $\text{corr}_\text{NeRF}$.
      By minimizing the discrepancy between $\text{corr}_\text{NeRF}$ and $\text{corr}_\text{diff}$, the pseudo ground-truth, we correct NeRF's 3D infidelities in the NeRF depths.
}
\vspace{-5.0mm}
\label{fig:main_overview}
\end{figure*}

\section{Method}
\label{method}

\smallbreak
\noindent
\textbf{Motivation and overview.}
Employing multi-view diffusion as a prior for 3D generation enhances the consistency of NeRF-rendered views, mitigating common issues such as the Janus face problem or content drift.
However, these priors are still confined to 2D space, which often results in errors in the geometric fidelity of the 3D output. While 2D rendered views may appear realistic, the 3D geometry can be flawed, exhibiting issues such as unnatural concavities or missing surfaces, as shown in~\cref{fig:3d_infidelity}.

We introduce \methodName, a novel method designed to improve the 3D fidelity of zero-shot text-to-3D outputs, using cross-view correspondences derived from the multi-view diffusion prior.
Our approach involves optimizing a NeRF model through both SDS and cross-view correspondence losses.
As the SDS loss adopts the conventional form as seen in ~\cref{eqn:multiview_sds}, we detail the correspondence loss in this section.
We generate two adjacent sets of NeRF-rendered views with minimally separated camera positions in azimuth (\cref{subsec:adjacent_rendering}), extract 2D features from the diffusion model (\cref{subsec:free_feature}), compute cross-view correspondences between adjacent rendered views (\cref{subsec:free_correspondence}), and use them to correct NeRF geometry via the cross-view correspondence loss (\cref{subsec:correspondence_loss}). 
The NeRF optimization using the SDS and cross-view correspondence losses are detailed in~\cref{subsec:nerf_optimization}.
~\cref{fig:main_overview} illustrates an overview of \methodName.

\subsection{Adjacent multi-view NeRF rendering}
\label{subsec:adjacent_rendering}
Our approach utilizes a multi-view diffusion model \( \epsilon_\theta \) which can concurrently generate \( N \) images \( \{x_t^{(i)}\}_{i=1}^N \), where each image is associated with a distinct viewpoint derived from the camera parameters and the given textual prompt \( y \).
These images represent a range of equispaced azimuth angles, capturing different perspectives of the same scene.
To effectively optimize the NeRF model \( \phi \), we would follow~\cref{eqn:multiview_sds} to render \( N \) views \( \{g(\phi,c_i)\}_{i=1}^N \), where \( c_i \) defines the camera parameters corresponding to the \( i \)-th view, and \( g \) is the NeRF rendering function dependent on the parameters of \( \phi \).
Through this process, the NeRF model is supervised to produce images that are consistent with the specified perspectives of \( c_i \), aligning the NeRF-rendered views with the diffusion model's predictions.

Due to the large azimuthal distances between each of the
$N$ views, there is limited viewpoint overlap between adjacent views, making direct correspondence computation challenging and prone to error.
To address this, we render two interlinked sets of \( N \) views, \( V_1 \) and \( V_2 \), ensuring that each view in \( V_1 \) has an adjacent view in \( V_2 \), thereby minimizing azimuthal separation.
The azimuth angles for the two sets are articulated as \( \{ \alpha_i \}_{i=1}^{N} \) and \( \{ \beta_i \}_{i=1}^{N} \), where \( \beta_i \) is defined as \( \alpha_i + \Delta \alpha \), with \( \Delta \alpha \) being a small, predetermined angular increment.
This approach simplifies the computation of correspondences by providing more overlapping fields of view,
thus ensuring a robust set of correspondences for subsequent optimization processes.

\subsection{Annotation-free feature extraction}
\label{subsec:free_feature}

We take advantage of the U-Net architecture within our multi-view diffusion model, which is adept at generating \( N \) synchronized views. 
In the optimization of 3D representations for text-to-3D generation, we add Gaussian noise $\eta$ to the NeRF-rendered views, modulated by a timestep $t$, creating noisy images $\tilde{v}_t^{(i)} = v_t^{(i)} + \sqrt{\alpha_t}\eta$, with $\eta$ distributed as $\mathcal{N}(0, \mathbf{I})$, and \( \alpha_t \) being a variance schedule function of the timestep \( t \), which controls the noise level. 
The diffusion model \( \epsilon_\theta \) then predicts the noise component as:
\[
\hat{\eta}^{(i)} = \epsilon_\theta(\tilde{v}_t^{(i)}; y, c_i, t)
\]
During this predictive step, we extract intermediate features \( \{ f_{l}^{(i)} \} \) from the U-Net's upsampling layers \( l \).
We build on existing studies that demonstrate the robustness of multi-layer features~\cite{kim2022transformatcher, luo2023dhf} to extract intermediate features across multiple layers.
These features are expressed as:
\begin{equation}
f_{l}^{(i)} = U_l(\tilde{v}_t^{(i)}; \theta_{l})
\end{equation}
where \( U_l \) is the upsampling function at layer \( l \), and \( \theta_{l} \) are the learned parameters specific to that layer.
This process yields a comprehensive set of features without additional training or explicit feature extraction algorithms.
Previous studies~\cite{tang2023dift, li2023sd4match} show that these diffusion U-Net features are surprisingly informative and discriminative, enabling the establishment of robust image correspondences.

\subsection{Cross-view correspondence computation}
\label{subsec:free_correspondence}

After obtaining the multi-view features for \( 2N \) views, \( \{ f_{l}^{(i)} \} \) and \( \{ f_{l}^{(i+N)} \} \) for \( i=1 \) to \( N \), we compute the correspondences between each pair of adjacent views, yielding \( N \) sets of adjacent-view correspondences. 
The feature maps extracted from the diffusion U-Net possess varying spatial dimensions across different layers, and are interpolated to a common resolution \( H' \times W' \), as follows:
\begin{equation}
f_{l}^{'(i)} = \mathcal{B}(f_{l}^{(i)}, H', W'), \quad f_{l}^{'(i+N)} = \mathcal{B}(f_{l}^{(i+N)}, H', W')
\end{equation}
where \( \mathcal{B} \) represents the bilinear interpolation function.

Prior to computing the correlation map, the feature maps are normalized to ensure comparability.
The correlation map \( C_{l}^{(i)} \) at each feature level \( l \) is computed to encapsulate pairwise similarity across all spatial positions, resulting in a 4D tensor with dimensions \( H' \times W' \times H' \times W' \).
The element \( C_{l}^{(i)}(p, q, r, s) \) represents the L2 distance between the vectors at positions \( (p, q) \) and \( (r, s) \):
\begin{equation}
C_{l}^{(i)}(p, q, r, s) = \frac{f_{l}^{'(i)}(p, q) \cdot f_{l}^{'(i+N)}(r, s)}{\left\| f_{l}^{'(i)}(p, q) \right\|_2 \left\| f_{l}^{'(i+N)}(r, s) \right\|_2}
\end{equation}
Subsequently, we aggregate the correlation maps from all feature levels to form the cumulative correspondence map \( \mathcal{C}^{(i)} \) for each view \( i \), defined by the sum:
\begin{equation}
\mathcal{C}^{(i)} = \sum_{l} C_{l}^{(i)}
\end{equation}
This 4D correlation map \( \mathcal{C}^{(i)} \) integrates the feature-level similarities into a singular comprehensive map~\cite{kim2024hccnet}.

For each spatial location \( (p, q) \), correspondences are determined by identifying the position with the highest value in \( \mathcal{C}^{(i)} \), signifying the nearest neighbor:
\begin{equation}
\text{corr}(p, q) = \underset{r, s}{\arg\max} \ \mathcal{C}^{(i)}(p, q, r, s)
\label{eqn:argmax_corrmap}
\end{equation}
where \( \text{corr}(p, q) \) designates the corresponding spatial location in the adjacent view for the point at \( (p, q) \). 
This dense correspondence field between each pair of adjacent views serves as a 3D geometric prior, which is instrumental in supervising the NeRF model for improved 3D fidelity.

As we know the ground-truth camera parameters for each rendered view, we can filter out implausible correspondences, adhering to constraints like the epipolar constraint.
We guide the readers to the supplementary for the details of correspondence post-processing.

\subsection{Cross-view correspondence loss}
\label{subsec:correspondence_loss}

Having established \( N \) sets of correspondences between adjacent NeRF-rendered image pairs, we leverage the depth information provided by the NeRF rendering process alongside the known camera parameters to reproject points from one view to the corresponding points in the adjacent view through a reprojection function denoted as \( \pi \).
For each pair of adjacent images, we now possess two distinct sets of correspondences: one derived from the diffusion features, \( \text{corr}_{\text{diff}} \), and the other obtained via reprojection using camera parameters and NeRF-rendered depths, \( \text{corr}_{\text{NeRF}} \). 
The reprojection function is defined as:
\begin{equation}
\text{corr}_{\text{NeRF}}(p) = \pi(\text{depth}_\phi(p), c, p)
\end{equation}
where \( \text{depth}_\phi(p) \) is the depth value at pixel \( p \), and \( c \) represents the camera parameters.

Our assumption posits that diffusion features are both informative and discriminative, yielding correspondences that not only associate semantically similar features but also adhere to geometric consistency, aligning with human perceptual reasoning. 
To enforce this assumption, we take inspiration from SPARF~\cite{truong2023sparf} to formulate a cross-view correspondence loss, which penalizes the NeRF-reprojected correspondences when they diverge from the diffusion feature correspondences, \ie, incoherent to common sense:
\begin{equation}
\mathcal{L}_{\text{corr}} = \sum_{p} \omega(p) \cdot \text{Huber}(\text{corr}_{\text{diff}}(p), \text{corr}_{\text{NeRF}}(p))
\end{equation}
Here, \( \text{Huber}(\cdot) \) represents the Huber loss function~\cite{hastie2009elements}, and \( \omega(p) \) is a weighting factor proportional to the similarity value at position \( p \) in the correlation map, enhancing the influence of high-confidence correspondences. 
This loss function serves to align the NeRF model's depth predictions with the geometrically and semantically robust correspondences derived from diffusion features, thereby correcting infidelities in the NeRF-rendered depths and enhancing the model's coherence to common sense.

\subsection{NeRF optimization}
\label{subsec:nerf_optimization}
In optimizing the NeRF model, we consider two distinct objectives: \( \mathcal{L}_{\text{SDS}} \), which ensures that NeRF-rendered views are consistent with the pre-trained diffusion model, and \( \mathcal{L}_{\text{corr}} \), which improves the 3D fidelity of the NeRF-inferred geometry.
Given the potential for these objectives to conflict—wherein the 3D geometry updates from \( \mathcal{L}_{\text{corr}} \) may not align with updates steered by \( \mathcal{L}_{\text{SDS}} \)—we employ an alternating optimization strategy to mitigate conflict between the objectives. 
This strategy is particularly pertinent as features conducive to accurate correspondence are typically extracted at lower timesteps (e.g., \( t=0 \)), while \( \mathcal{L}_{\text{SDS}} \) benefits from a wide range of sampled timesteps during the 3D optimization process~\cite{poole2022dreamfusion, shi2023mvdream}.

We define the total number of optimization iterations as \( T \), with a predefined range \( [t_{\text{start}}, t_{\text{end}}] \) within which \( \mathcal{L}_{\text{corr}} \) is active. 
The SDS loss is applied to NeRF at every iteration by default.
However, for iterations \( t \) such that \( t_{\text{start}} \leq t \leq t_{\text{end}} \) and \( t \) is even ($t \% 2 = 0$), we alternate to apply \( \mathcal{L}_{\text{corr}} \) without \( \mathcal{L}_{\text{SDS}} \).
This alternating approach leverages the strengths of both losses, facilitating a balanced optimization that enhances both the visual coherence and 3D geometric fidelity of the NeRF-rendered scenes.

\begin{figure*}[ht]
    \begin{center}
        \includegraphics[width=0.9\linewidth]{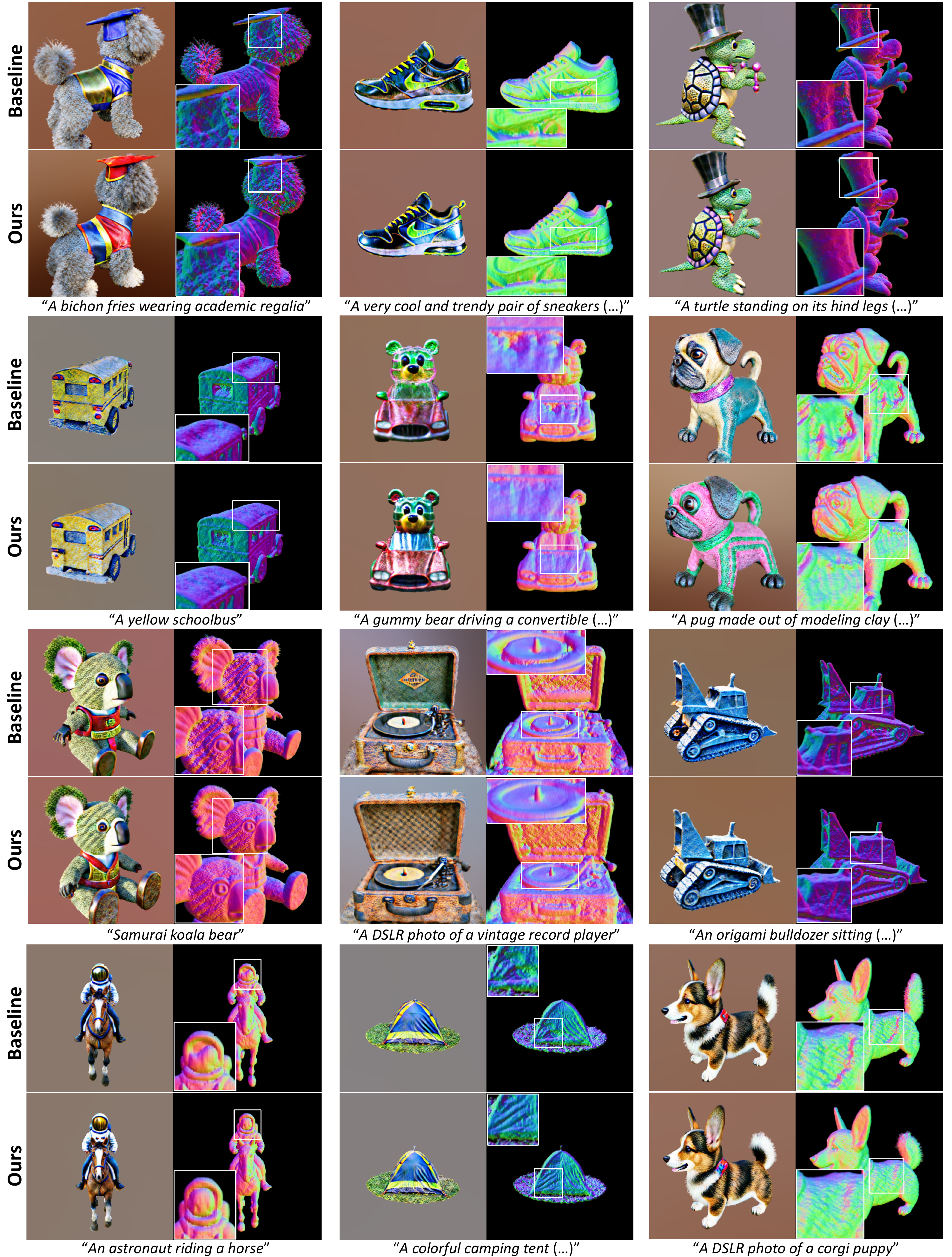}
    \end{center}
    \vspace{-5.0mm}
      \caption{\textbf{Qualitative results of our \methodName across various prompts}. 
      It can be seen that \methodName yields substantially improved 3D fidelity across various prompts.
      The 3D infidelities from the baseline (MVDream~\cite{shi2023mvdream}) are highlighted and zoomed in white squares. 
      Best viewed on electronics, zoom in for better visualization.
}
\label{fig:main_qual}
\end{figure*}

\section{Experiment}
\label{sec:experiment}

\subsection{Implementation details}
We implement \methodName on top of the open-sourced multi-view diffusion model MVDream~\cite{shi2023mvdream}, which implements a zero-shot text-to-3D pipeline on top of the threestudio~\cite{threestudio2023} library.
We use final image dimensions of 128$\times$128 for NeRF-rendered views for improved latency and memory overhead\footnote{Stable Diffusion v2.1 base model~\cite{rombach2022high} generates images at resolutions of 512$\times$512.
MVDream~\cite{shi2023mvdream} uses a reduced image size of 256$\times$256 when finetuning the Stable Diffusion model for multi-view image generation, and also when rendering NerF-rendered views for text-to-3D.}.
We noticed that the quality of text-to-3D using MVDream is maintained at image dimensions of 128$\times$128; and in the presence of 3D infidelities, the errors are also consistent at these image resolutions.
We illustrate qualitative evidence for this in the supplementary.%

For our NeRF, we use the implicit-volume implementation in the Nerfacc library~\cite{li2023nerfacc}.
We use $t_{\text{start}}=3000$ and $t_{\text{end}}=7000$, leading to 2000 iterations of correspondence loss supervision without SDS supervision.
To compensate for this, we optimize our NeRF model for a total number of iterations $T=12000$\footnote{which is 2000 higher than the number of iterations used in MVDream.}.
We optimize the NeRF model using an AdamW optimizer~\cite{loshchilov2018decoupled} with a constant learning rate of 0.01.
$L_{SDS}$ and $L_{\text{corr}}$ are weighted at 1.0 and 1,000 respectively.
For adjacent multi-view NeRF rendering, we uniformly sample from $[ 10^{\circ},30^{\circ}]$ for $\Delta \alpha$, and ensure that the adjacent views are added with the same noise, modulated with $t=0$\footnote{
Inspired from existing work on diffusion-based image matching~\cite{tang2023dift,li2023sd4match} which show that $t=0$ gives the most robust features.}.
We use the output feature maps from the 6th and 9th upsampling layers of the UNet to compute the correlation map.
The NeRF optimization with normal rendering takes about 2 hours on a Tesla V100 GPU.
\subsection{Qualitative results}
We present comparative qualitative results between \methodName and MVDream~\cite{shi2023mvdream}.
Other 2D lifting methods that rely on single-view diffusion model priors~\cite{poole2022dreamfusion, lin2023magic3d} suffer from unresolved multi-view consistency issues, \ie, the Janus face and the content drift problems.
This overwhelms the 3D infidelities, making in inappropriate to qualitatively compare against such methods (\cref{sec:supp_other_models}).
The results are shown in~\cref{fig:main_qual}, where it can be seen that \methodName can visibly remove 3D infidelities \ie, concavities or missing surfaces, across various prompts, improving the 3D fidelity of NeRF-rendered geometry.

As our proposed cross-view correspondence affects the NeRF model directly via the NeRF-rendered depths, it can be seen that the output appearance \ie, size, color or the overall appearance, of the output itself may also differ compared to using the SDS loss alone.
For example, the 3D outputs of the prompt "\textit{A bichon fries wearing academic regalia}" are wearing differently coloured regalias, and the 3D outputs of the prompt "\textit{A pug made out of modeling clay}" exhibit different colours as well.
Nonetheless, the output when using our method is still coherent with the input prompt, but with improved 3D fidelity.

\subsection{Comparative analysis}
In this section, we perform analytical experiments to qualitatively evidence the design choices of our approach.

\begin{figure}[h]
    \begin{center}
        \includegraphics[width=\linewidth]{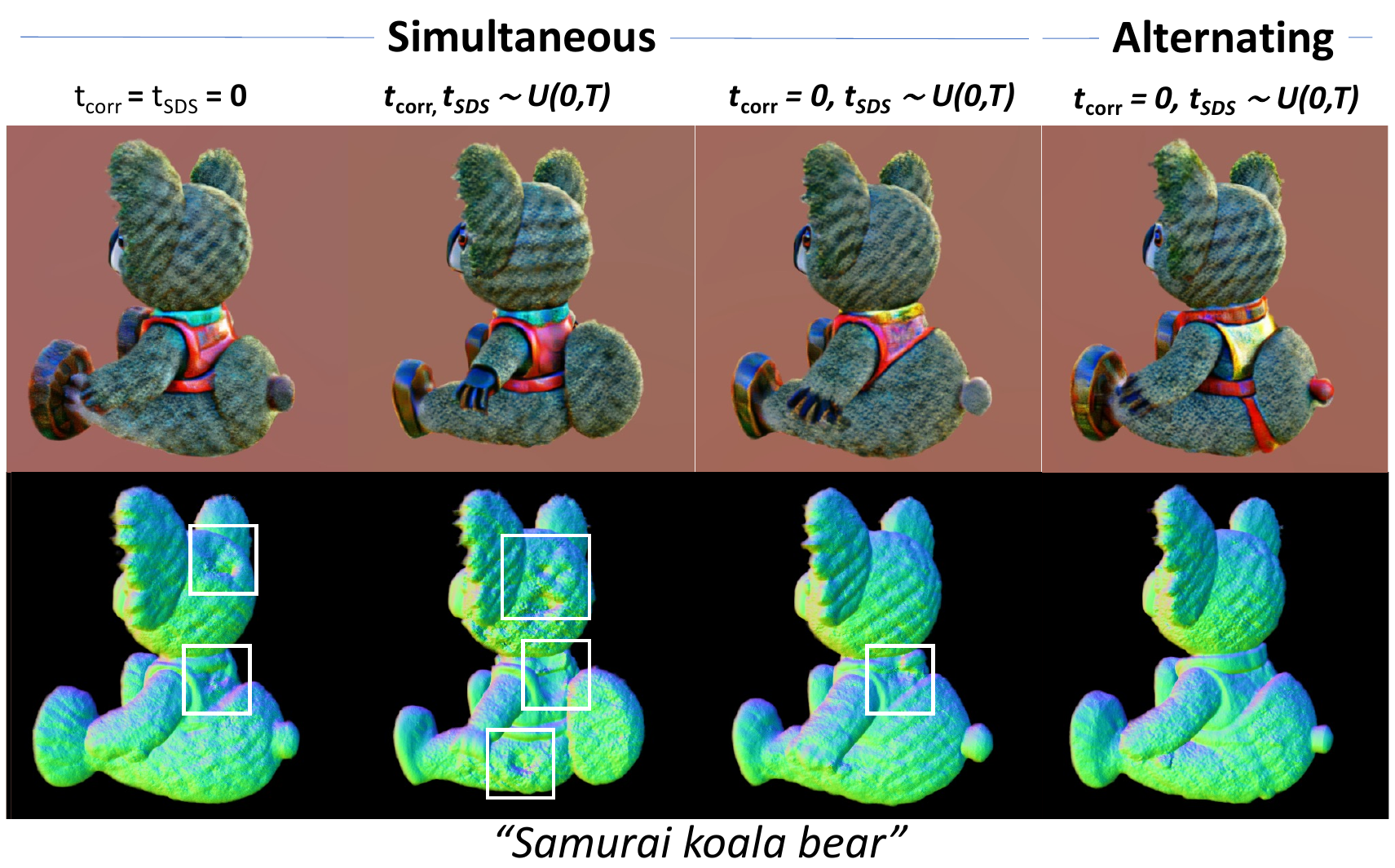}
    \end{center}
    \vspace{-5.0mm}
      \caption{\textbf{Analysis of alternating supervision.} 
      Noticeable 3D inaccuracies are marked with white squares.
      Our alternating supervision approach demonstrates superior qualitative outcomes. 
}
\vspace{-4.0mm}
\label{fig:ablation_alternation}
\end{figure}

\smallbreak
\noindent
\textbf{Analysis on alternating supervision of $\mathcal{L}_{\text{SDS}}$ and $\mathcal{L}_{\text{corr}}$.}
We compare the scheme of supervising NeRF with alternating \( \mathcal{L}_{\text{SDS}} \) and $\mathcal{L}_{\text{corr}}$ to non-alternating (simultaneous) alternatives.
Under the simultaneous setting, we either (1) fix the timestep $t$ to be always randomly sampled (as done for \( \mathcal{L}_{\text{SDS}} \)), or (2) always set at $t=0$ (as done for $\mathcal{L}_{\text{corr}}$), or (3) randomly sample the timestep $t$ for \( \mathcal{L}_{\text{SDS}} \), and use $t=0$ for $\mathcal{L}_{\text{corr}}$ when modulating the diffusion model.
Note that the (3) results in increased computation costs as the consequence of multiple forwards of the diffusion model for the same input.
We illustrate the results in~\cref{fig:ablation_alternation}, where it can be seen that our alternating scheme yields the best results.

\begin{figure}[h]
    \begin{center}
        \includegraphics[width=\linewidth]{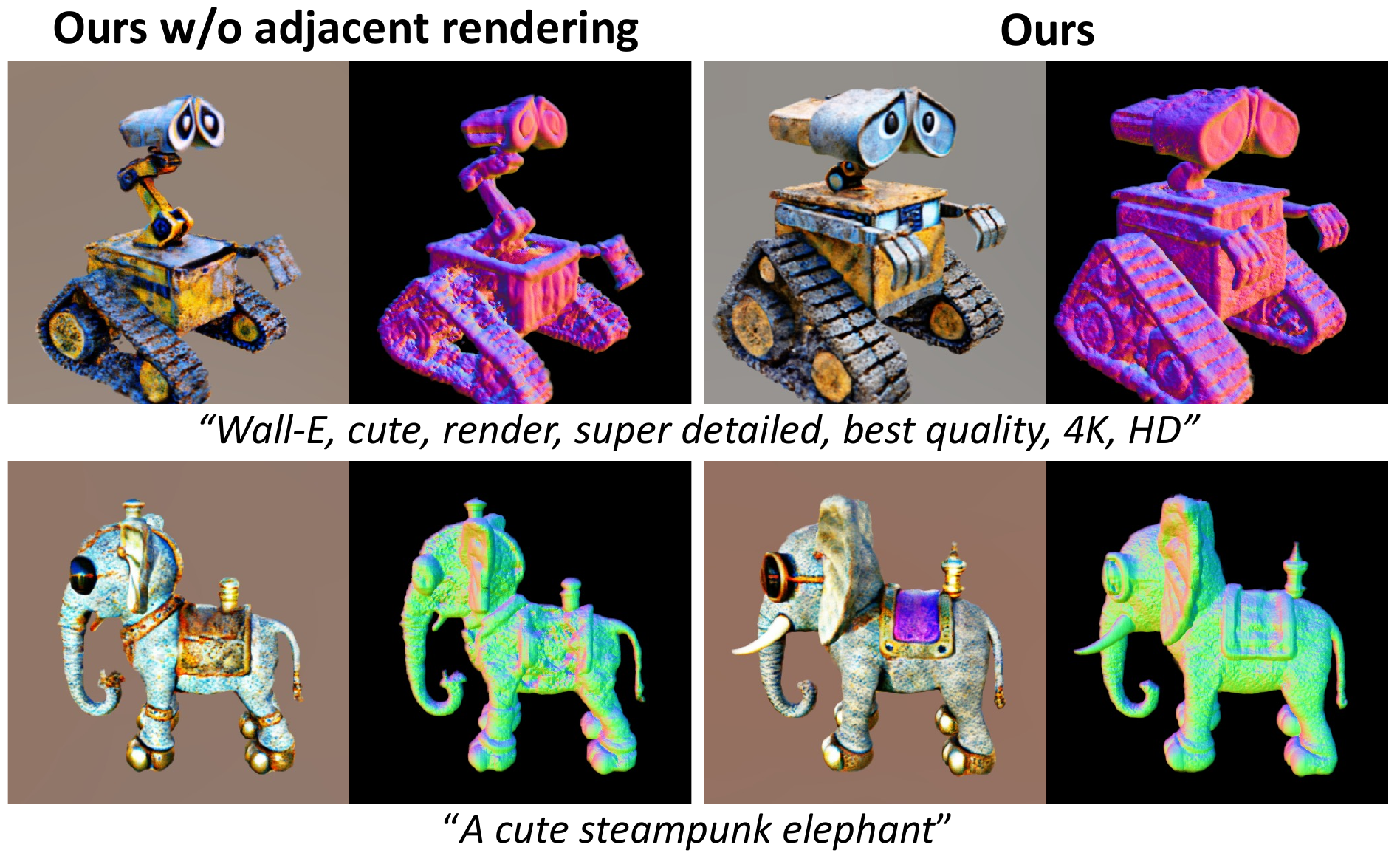}
    \end{center}
    \vspace{-5.0mm}
      \caption{\textbf{Ablation of adjacent multi-view rendering.} 
      Without adjacent rendering, we establish correspondences between adjacent views within only a single set of rendered views.
      This results in a very small overlapping region, leading to erroneous cross-view correspondence, and consequently worsened output.
}
\vspace{-5.0mm}
\label{fig:ablation_adjacent}
\end{figure}

\smallbreak
\noindent
\textbf{Ablation on adjacent multi-view rendering.}
We perform an ablation on adjacent multi-view rendering, and show the results in~\cref{fig:ablation_adjacent}.
It clearly shows that our current scheme of adjacent multi-view rendering yields much better results.
As explained in~\cref{subsec:adjacent_rendering}, the azimuthal distance between adjacent views within a single set of multi-view renderings would be too large, \ie, lower overlap region between viewpoints, making it challenging to establish dense, robust correspondences for appropriate supervision.
\subsection{User study}
\begin{table}[ht]
\centering
\begin{tabular}{lcc}
\hline
Method & User preference \% \\
\hline
MVDream~\cite{shi2023mvdream} & 30.4 \\
\methodName (ours) & \textbf{69.6} \\
\hline
\end{tabular}
\caption{\textbf{User study.} Users were asked to pick their preference based on perceived 3D fidelity and overall quality.
Our method was selected more than twice compared to the baseline~\cite{shi2023mvdream}.
}
\label{tab:user_study}
\end{table}

Due to the absence of ground-truth 3D scenes corresponding to text prompts, it is difficult to conduct a quantitative evaluation on the 3D fidelity of the text-to-3D outputs.
Instead, we perform a user study on generated models from 40 non-cherry-picked prompts, where each user is asked to select the preferred 3D model in terms of the 3D fidelity \ie, coherence to the 2D images, and the overall quality of the output.
767 responses from 25 participants were collected, and the results are shown in Table~\ref{tab:user_study}.
It can be seen that \methodName was selected to be more favorable more than twice compared to MVDream alone.

\subsection{Drawbacks and failure cases.}
\begin{figure}[h]
    \begin{center}
        \includegraphics[width=0.75\linewidth]{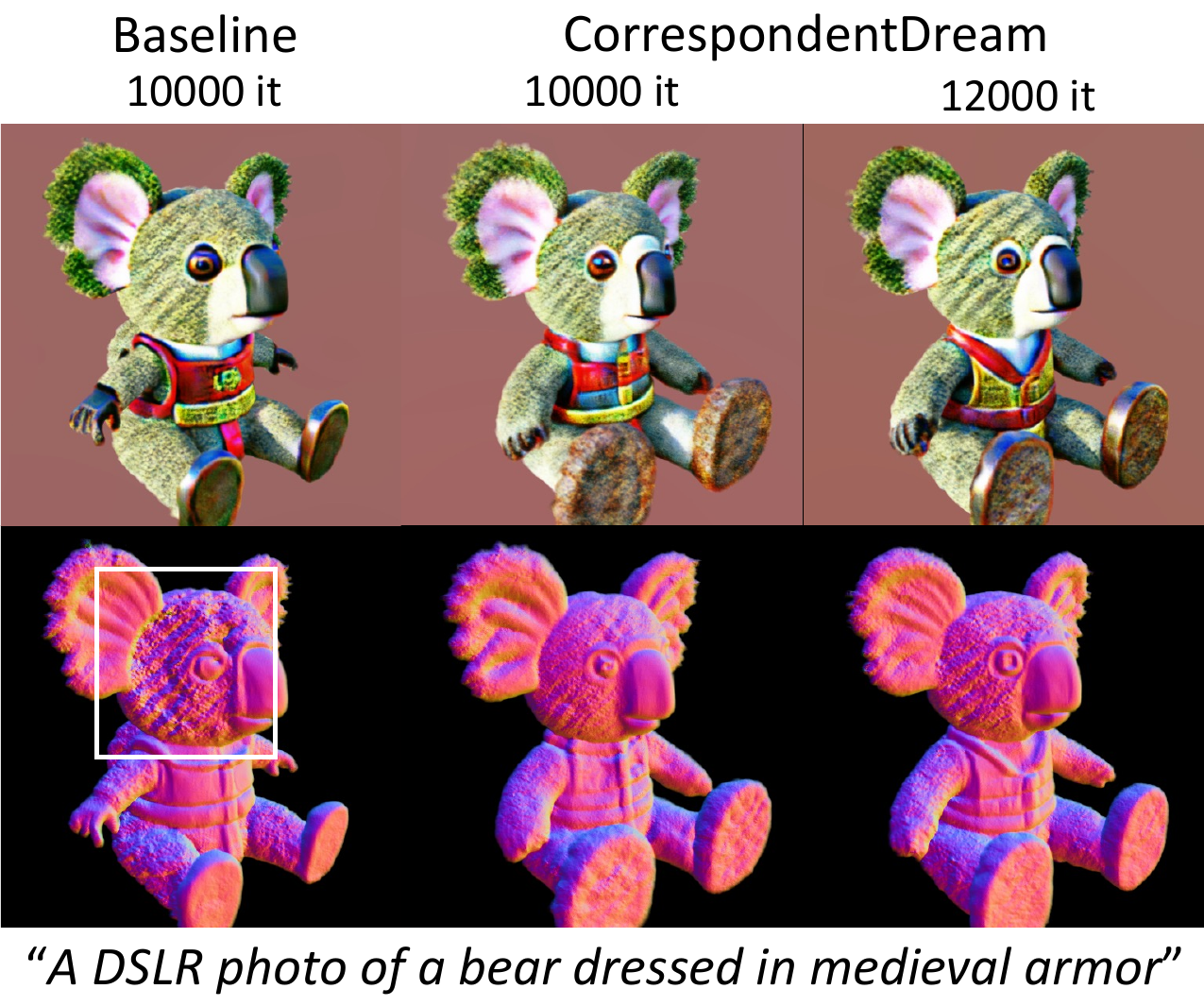}
    \end{center}
    \vspace{-5.0mm}
      \caption{\textbf{Varying iterations.} 
      Even with the same number of iterations as the baseline~\cite{shi2023mvdream}, \ie, lower number of effective SDS optimization steps, \methodName still improves the 3D fidelity of the output while achieving high-quality colour and texture.
}
\label{fig:ablation_iteration}
\end{figure}

\begin{figure}[h]
    \begin{center}
        \includegraphics[width=\linewidth]{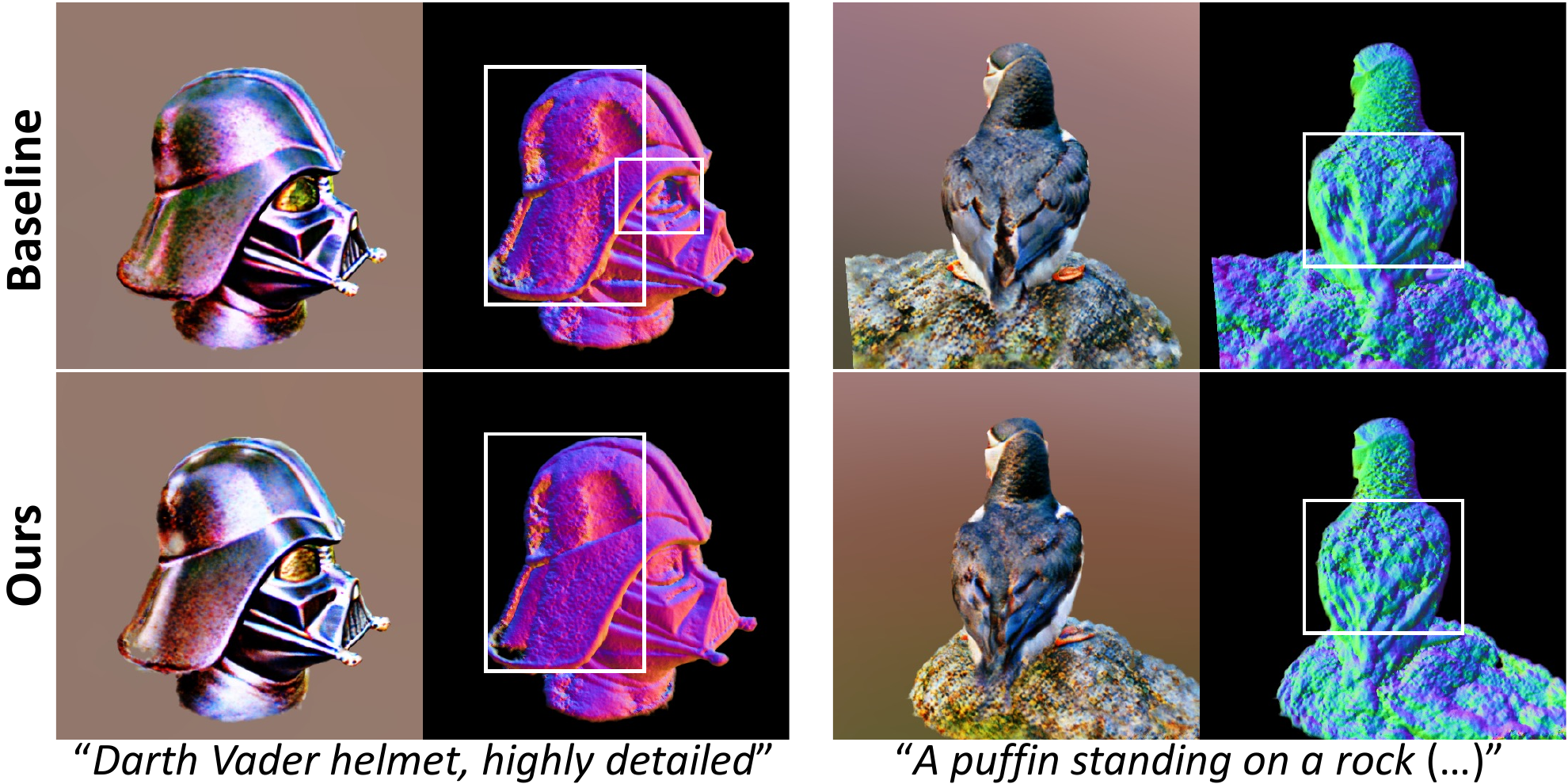}
    \end{center}
    \vspace{-5.0mm}
      \caption{\textbf{Failure cases.} 
      In the presence of shiny homogeneous surfaces (helmet, left), or many repetitive patterns (feathers, right), our method occasionally falls short at correcting the 3D infidelity.
}
\vspace{-2.0mm}
\label{fig:failure_case}
\end{figure}

The main drawback of our method is the increase in optimization iterations due to the alternating supervision. 
However, we note that we used a larger number of iterations to keep the number of SDS-supervised iterations the same with the baseline for a fair comparison.
Figure~\ref{fig:ablation_iteration} shows the result when we use the same number of iterations as the baseline.
\methodName still shows high-quality colour and texture albeit improved fidelity\footnote{This \textit{may} result in a different appearance depending on the prompt, as the effective number of iterations for SDS optimization is decreased.}.

\methodName shows to often fail in cases where 
there are shiny homogeneous surfaces or repeated patterns within the image as shown in Figure~\ref{fig:failure_case}.
We conjecture this is because such cases may be challenging for diffusion features to yield robust and dense correspondences between the rendered views, being unable to provide sufficiently informative 3D prior during the NeRF optimization.

\section{Conclusion}
\label{conclusion}

We have presented \methodName, a novel method that leverages annotation-free, cross-view correspondences computed from diffusion features to additionally supervise the 3D representation in zero-shot text-to-3D models for improved 3D fidelity. 
By formulating the cross-view correspondence loss computed using the NeRF-reprojected pixels and the cross-view correspondences, we can correct the geometric flaws in NeRF depths caused by the ambiguities that cannot be handled by multi-view 2D image priors alone.
Notably, this does not require any additional explicit priors or off-the-shelf modules.
We demonstrate the efficacy of our approach via qualitative results and a user study on a large collection of varying text prompts.
Our work aims to shed light onto the neglected issue of 3D geometric fidelity of diffusion-guided text-to-3D models, paving the way for enhanced applicability to practical scenarios.

\vspace{2mm}
\noindent
\textbf{Acknowledgement.} 
This work was done while Seungwook Kim was an intern at ByteDance. 
Seungwook Kim was supported by the Hyundai-Motor Chung Mong-koo Foundation.
This work was also supported by IITP grants (No.2021-0-02068: AI Innovation Hub (90\%), No.2019-0-01906: Artificial Intelligence Graduate School Program at POSTECH (10\%)) funded by the Korea government (MSIT).

\clearpage
\appendix
\setcounter{table}{0}
\renewcommand{\thetable}{A\arabic{table}}%
\setcounter{figure}{0}
\renewcommand{\thefigure}{A\arabic{figure}}%
\maketitlesupplementary

In this supplementary material, we provide additional details and qualitative results of \methodName which were not included in the main paper due to spatial constraints.
We detail the refining and post-processing of correspondences in~\cref{sec:supp_correspondence_postprocessing}, and explain additional implementation details of \methodName in~\cref{sec:supp_implementation_details}.
In~\cref{sec:supp_cfg_scheduling}, we elaborate on the effect the value of CFG ($\omega$) has on the generated 3D object, and devise a CFG scheduling scheme to yield more satisfying results.
We visualize the cross-view correspondences in~\cref{sec:supp_correspondence_visualization}.
We analyze the results when using cross-view correspondence loss as a pre-processing or post-processing method in~\cref{sec:supp_prepostprocessing}.
We demonstrate that even at reduced image resolutions used in \methodName, the quality and 3D infidelities of the outputs are maintained in~\cref{sec:supp_image_resolution}.
We visualize the intermediate rendered outputs of \methodName in comparison to the baseline (MVDream~\cite{shi2023mvdream}) in~\cref{sec:supp_progressive_visualization}.
We elaborate on the applicability of \methodName on other zero-shot text-to-3D generation methods in~\cref{sec:supp_other_models}.
We outline the computation cost of \methodName, and evidence that the improved 3D fidelity does not come from the additional computation in~\cref{sec:computational_cost}.
We demonstrate that using off-the-shelf image matchers give dissatisfactory results, and substantiate the use of annotation-free cross-view correspondences in~\cref{sec:rebuttal_offtheshelf_matcher}.
Finally, we present some example text prompts used for our experiments in~\cref{sec:supp_example_prompts}.

\section{Correspondence post-processing details}
\label{sec:supp_correspondence_postprocessing}

In~\cref{subsec:free_correspondence}, we mentioned that the correspondences are determined from the correlation map as shown in~\cref{eqn:argmax_corrmap}.
In this section, we describe the refining and post-processing steps which were applied to the correlation map and the sampled correspondences to yield a more robust set of correspondences that are aligned with human common sense.

\begin{figure}[h]
    \begin{center}
        \includegraphics[width=\linewidth]{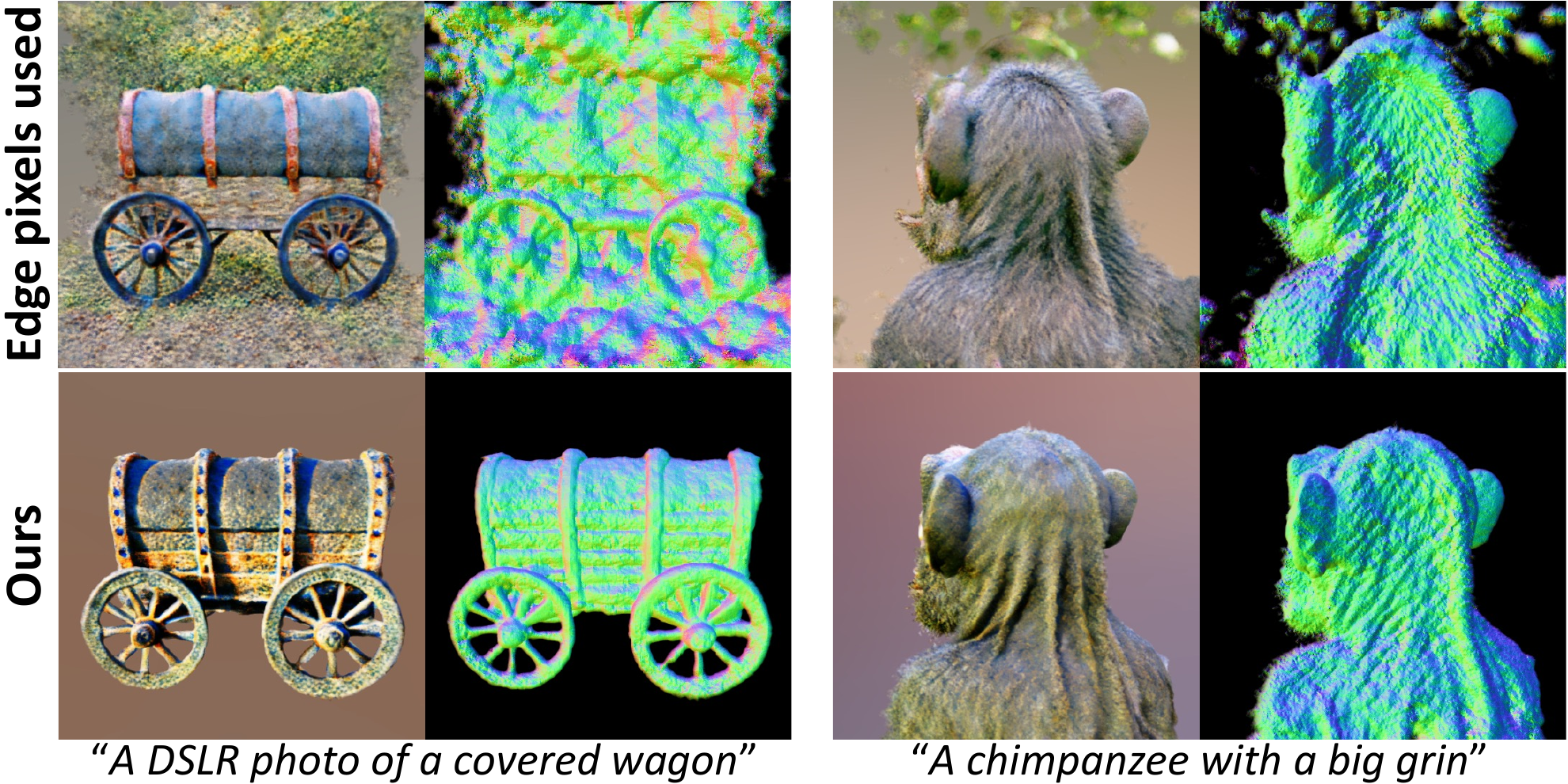}
    \end{center}
      \caption{\textbf{Ablation of \methodName opacity-based foreground edge discarding}. 
      It can be seen that not discarding the foreground edge pixels during the cross-view correspondence loss results in unwanted artifacts around the final output.
}
\label{fig:supp_opacity_artifact}
\end{figure}

\smallbreak
\noindent
\textbf{Motivation.}
As we do not have the ground-truth correspondences between rendered views, we couldn't directly evaluate if our cross-view correspondences align sufficiently well with human perception in the presence of 3D infidelities.
To this end, we devised an indirect protocol to ensure that our annotation-free cross-view correspondences are consistent with human perception.

We sampled 10 text-to-3D outputs which seems to have near-perfect 3D fidelity by human eyes.
This would mean that the NeRF depths are also consistent with human perception, allowing us to consider the NeRF reprojections as the pseudo-ground truth correspondences.
We experimented with various refinement / post-processing settings to maximize the precision and recall of our annotation-free cross-view correspondences with respect to the pseudo-ground truth correspondences.

\smallbreak
\noindent
\textbf{1. Filtering out by opacity.}
To ensure that we are establishing correspondences only between the foreground objects, \ie, not the background, we use the NeRF opacity values to filter out the background positions.
The opacity value ranges from 0 to 1, where 0 signifies no occupancy (background) and values closer to 1 signifies high occupancy (likely to be foreground). 
We can therefore disregard the background positions by filtering out pixel positions with opacity$=0$.

However, we noticed that using the edge pixels (\ie, neighboring a background pixel) of the foreground object results in unwanted artifacts near the edge of the object as exemplified in~\cref{fig:supp_opacity_artifact}. 
Considering that non-edge pixels of the foreground usually have a opacity value of $>$0.99, we additionally discard the edge pixels from the cross-view correspondence loss computation by performing 2D average pooling on the opacity map, and disregarding the pixels with opacity values less than the threshold value of 0.99.
This step is carried out for both the source and target images, where the predicted target pixel should also be within the non-edge foreground pixels of the rendered view.

\smallbreak
\noindent
\textbf{2. Soft mutual nearest neighbours filtering.}
After we compute the 4D correlation map between adjacently rendered views as explained in~\cref{subsec:free_correspondence} of the main paper, we perform a soft mutual nearest neighbour filtering as proposed in NCNet~\cite{rocco2018neighbourhood} to facilitate the reciprocity constraint on matches.

For self-containedness, we provide the details of this approach in the following.
Given a correlation map $C^{(i)}$, we perform a soft mutual nearest neighbour filtering $M(\cdot)$ to yield a refined feature map $\hat{\mathcal{C}}^{(i)} = M(\mathcal{C}^{(i)})$, where $\hat{\mathcal{C}}^{(i)}(p,q,r,s) = r^{(i)}_{pqrs}r^{(i+N)}_{pqrs}\mathcal{C}(p,q,r,s)$.
$r^{(i)}_{pqrs}$ and $r^{(i+N)}_{pqrs}$ are ratios of the score of the particular match $\mathcal{C}^{(i)}(p,q,r,s)$ with the best scores along each pair of dimensions corresponding to the $i$-th and $(i+N)$th view respectively:
\begin{equation}
r^{(i)}_{pqrs} = \frac{\mathcal{C}^{(i)}(p,q,r,s)}{\text{max}_{ab}\mathcal{C}^{(i)}(a,b,r,s)}, r^{(i + N)}_{pqrs} = \frac{\mathcal{C}^{(i)}(p,q,r,s)}{\text{max}_{cd}\mathcal{C}^{(i)}(p,q,c,d)}.
\end{equation}

\smallbreak
\noindent
\textbf{3. 4D smoothing.}
We additionally perform a 4D smoothing operation on the correlation map.
This not only intends to smooth the 4D space of image correspondences, but also aims at disambiguating correspondences with the help of neighbouring matches.
This motivation is inspired by the NCNet~\cite{rocco2018neighbourhood} as well; we can assume correct matches to have a coherent set of supporting them in the 4D space, especially when the two rendered views depict the \textit{same} object but just from different azimuthal viewpoints.
Our 4D smoothing acts as a \textit{soft} spatial consensus constraint (4D convolution with $1/k^{3}$ uniform weight for each kernel position, instead of learnable weights) to avoid errors on ambiguous or textureless matches.

\smallbreak
\noindent
\textbf{4. Multi-layer features.}
We aim to leverage the multiple features that can be obtained along the varying depths of the upsampling layers of the diffusion U-Net. 
It has been empirically demonstrated in existing work on image correspondences~\cite{kim2022transformatcher, hedlin2023unsupervised, kim2024hccnet} that it is beneficial to leverage features from multiple layers, to exploit both the semantics/context and local patterns/geometries that are encoded in different layers.
Among the 12 upsampling layers of the diffusion U-Net in the multi-view diffusion model proposed by MVDream~\cite{shi2023mvdream}, we extract features from the 6th and 9th upsampling layers.
We tried with various other combinations, but using these two layers qualitatively gave satisfactory results with reasonable computation overhead.

\smallbreak
\noindent
\textbf{5. Epipolar constraint.}
For each of the rendered views from NeRF, we know the ground-truth camera parameters, specifically the extrinsic and intrinsic parameters.
Using these, we can accurately determine the epipolar line on the target view corresponding to any pixel on the source view.
The epipolar constraint states that the true corresponding point of a point from the source view must lie on the epipolar line.
Adhering to this constraint, we project all predicted target points to their respective epipolar lines.
To discard obviously wrong correspondences, we calculate the projection distance to the epipolar line in order to discard any correspondence whose projection distance to the epipolar line is larger than a pixel distance threshold $\tau_{\text{epi}}$.
We use $\tau_{\text{epi}} = 2$ in our settings.

\smallbreak
\noindent
\textbf{Out-of-bounds filtering.}
Finally, we filter out any predicted target points which fall out of the non-edge foreground pixels.
Also, in calculating the corr$_{\text{NeRF}}$ from NeRF reprojections, we filter out any reprojections if they fall out of the image bounds of $H', W'$.

\section{Additional implementation details}
\label{sec:supp_implementation_details}
In this section, we provide additional implementation details of \methodName.
It is noteworthy that \methodName was implemented largely on threestudio~\cite{threestudio2023} and MVDream~\cite{shi2023mvdream}, and the majority of settings detailed below can be manipulated on their codebase.

We follow the protocols outlined in MVDream~\cite{shi2023mvdream} to embed the camera embeddings together with the time embeddings as residuals, by adding them together prior to being input to the diffusion network.
To prevent the model from generating 3D models with low quality appearance and style, we further add a few fixed negative prompts during the SDS optimization, \eg, "blurry" or "low quality", following MVDream~\cite{shi2023mvdream}.

We sample $t\sim\mathcal{U}(t_{\text{min}},t_{\text{max}})$ where $t_{\text{min}}$ anneals from 0.98 to 0.02, and $t_{\text{max}}$ anneals from 0.98 to 0.5, both in a linear manner for 9600 iterations.
This is the same as MVDream~\cite{shi2023mvdream}, except that the annealing iterations were increased in proportion to the increased number of total iterations.
For the first 6000 iterations, we render views from the NeRF at image dimensions of 32$\times$32 at batch size of 8 (\ie, 2 sets of 4 views, total of 16 views rendered for cross-view correspondence loss).
After the 6000th iteration, we render views from the NeRF at image dimensions of 128$\times$128 (\ie, 1 set of 4 views, total of 8 views rendered for cross-view correspondence loss).
Beginning the NeRF optimization with rendered views at a lower resolution drastically reduces VRAM usage as empty space is pruned in early training.
We set $\omega=50$ for our class-free guidance.
We also devise a class-free guidance scheduling scheme for improved 3D generation quality in the next section (\cref{sec:supp_cfg_scheduling}).
We use a rescale factor of 0.5 for the CFG rescale trick~\cite{lin2023common}.
We turn on soft shading~\cite{lin2023magic3d} and point lighting~\cite{poole2022dreamfusion} to regularize the geometry.
We also use the modified version of the orientation loss~\cite{verbin2022refnerf} as in DreamFusion~\cite{poole2022dreamfusion}, to penalize normal vectors facing backwards away from the camera for the first 6000 iterations.
This orientation loss is weighted with a weight that scales linearly from 10 to 1000 until the 6000th iteration.
The background is replaced with 50\% chance to force the separation between the foreground and the background during the NeRF optimization.

\section{Class-free guidance scheduling}
\label{sec:supp_cfg_scheduling}
\begin{figure*}[ht]
    \begin{center}
        \includegraphics[width=0.9\linewidth]{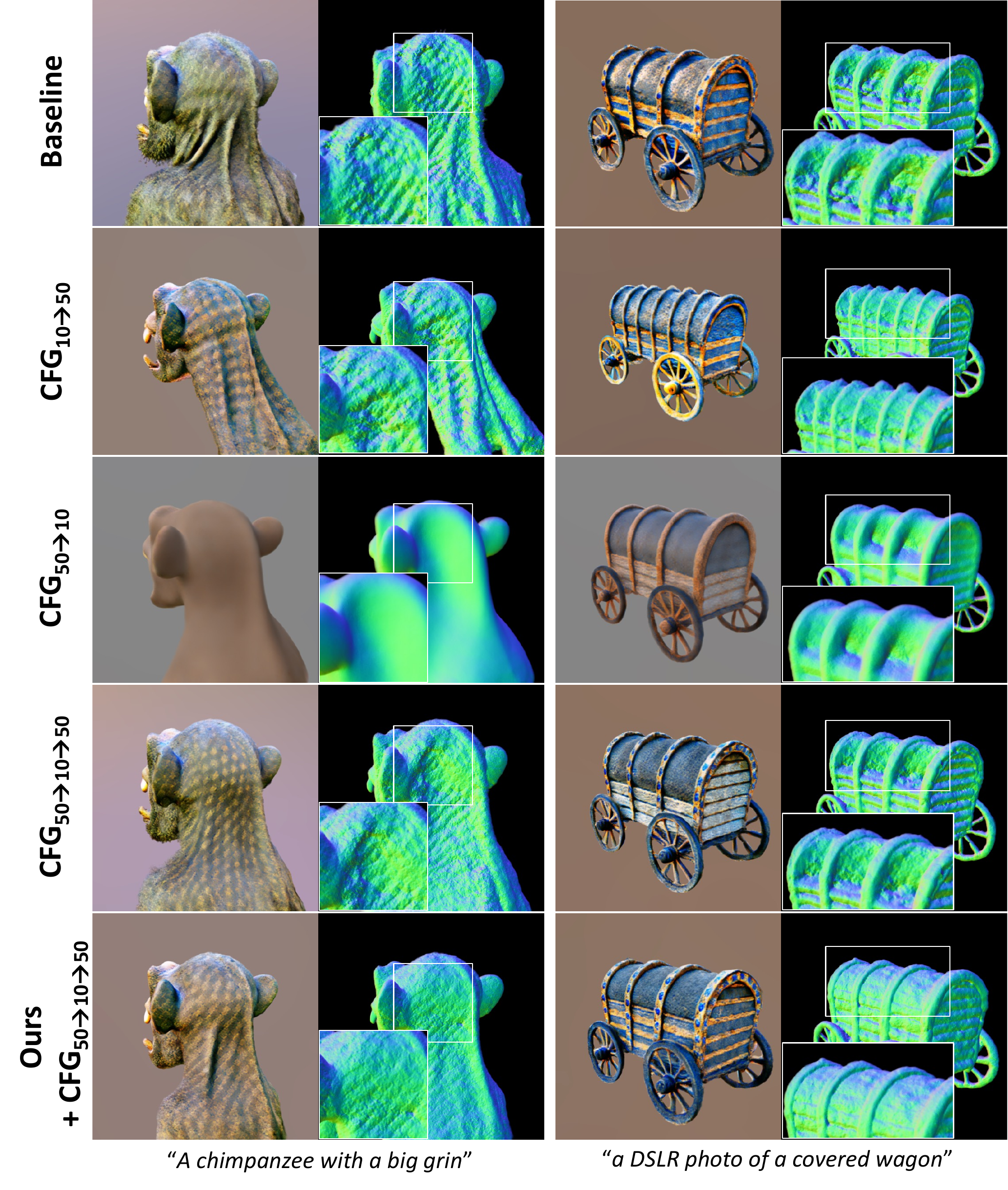}
    \end{center}
    \vspace{-5.0mm}
      \caption{\textbf{Visualization of CFG scheduling with and without $\mathcal{L}_{\text{corr}}$}. 
It can be seen that beginning at $\omega=10$ results in a smaller object, and ending with $\omega=10$ results in oversmoothed surfaces. 
While beginning at $\omega=50$ results in a larger object in comparison, ending with $\omega=50$ seems to end up with more severe cases of 3D infidelities.
CFG$_{50 \rightarrow 10 \rightarrow 50}$ exhibits a larger 3D object size, and an appropriate trade-off between the smoothness and detail of the generated 3D shape.
Nonetheless, CFG scheduling alone is insufficient to alleviate the 3D infidelities - it shows that incorporating \methodName together with CFG$_{50 \rightarrow 10 \rightarrow 50}$ exhibits the best qualitative results overall.
}
\label{fig:supp_cfg_scheduling}
\end{figure*}

Note that we are using a default CFG value of $\omega = 50$ otherwise mentioned in our qualitative visualizations.
Through various experiments, we noticed that the value of class-free guidance (CFG) $\omega$ has a dramatic effect on the level of details and smoothness of the rendered 3D object.
Specifically, we observed that using a large value for $\omega$ 1) results in a larger 3D object, 2) results in a higher level of details in the 3D object, but 3) has a larger risk of 3D infidelities.
On the other hand, we noticed that using a low value for $\omega$ 1) results in a smaller 3D object, 2) results in a smoother surface of 3D objects, but 3) holds a much lower level of details (overly smoothed) in the 3D object.
To this end, we experiment with various CFG scheduling schemes at an aim to devise a scheme to maximize the benefits of both high and low CFG values, while minimizing their downsides.
~\cref{fig:supp_cfg_scheduling} visualizes the results of different scheduling schemes we tried.
CFG$_{10 \rightarrow 50}$ denotes using $\omega = 10$ for first half, and $\omega = 50$ for the latter half of the 3D generation process.
CFG$_{50 \rightarrow 10}$ denotes using $\omega = 50$ for first half, and $\omega = 10$ for the latter half of the 3D generation process.
CFG$_{50 \rightarrow 10 \rightarrow 50}$ denotes using $\omega = 50$ for first third, and $\omega = 10$ for the second third, and $\omega = 50$ again for the last third of the 3D generation process.

It can be seen that beginning at $\omega=10$ results in a smaller object, and ending with $\omega=10$ results in oversmoothed surfaces. 
While beginning at $\omega=50$ results in a larger object in comparison, ending with $\omega=50$ seems to end up with more severe cases of 3D infidelities.
The qualitative results show that CFG$_{50 \rightarrow 10 \rightarrow 50}$ exhibits a larger 3D object size, and an appropriate trade-off between the smoothness and detail of the generated 3D shape.
Nonetheless, CFG scheduling alone is insufficient to alleviate the 3D infidelities - and it shows that incorporating CorrespondentDream together with CFG$_{50 \rightarrow 10 \rightarrow 50}$ exhibits the best qualitative results overall.

\section{Correspondence visualization}
\label{sec:supp_correspondence_visualization}
\begin{figure*}[ht]
    \begin{center}
        \includegraphics[width=0.9\linewidth]{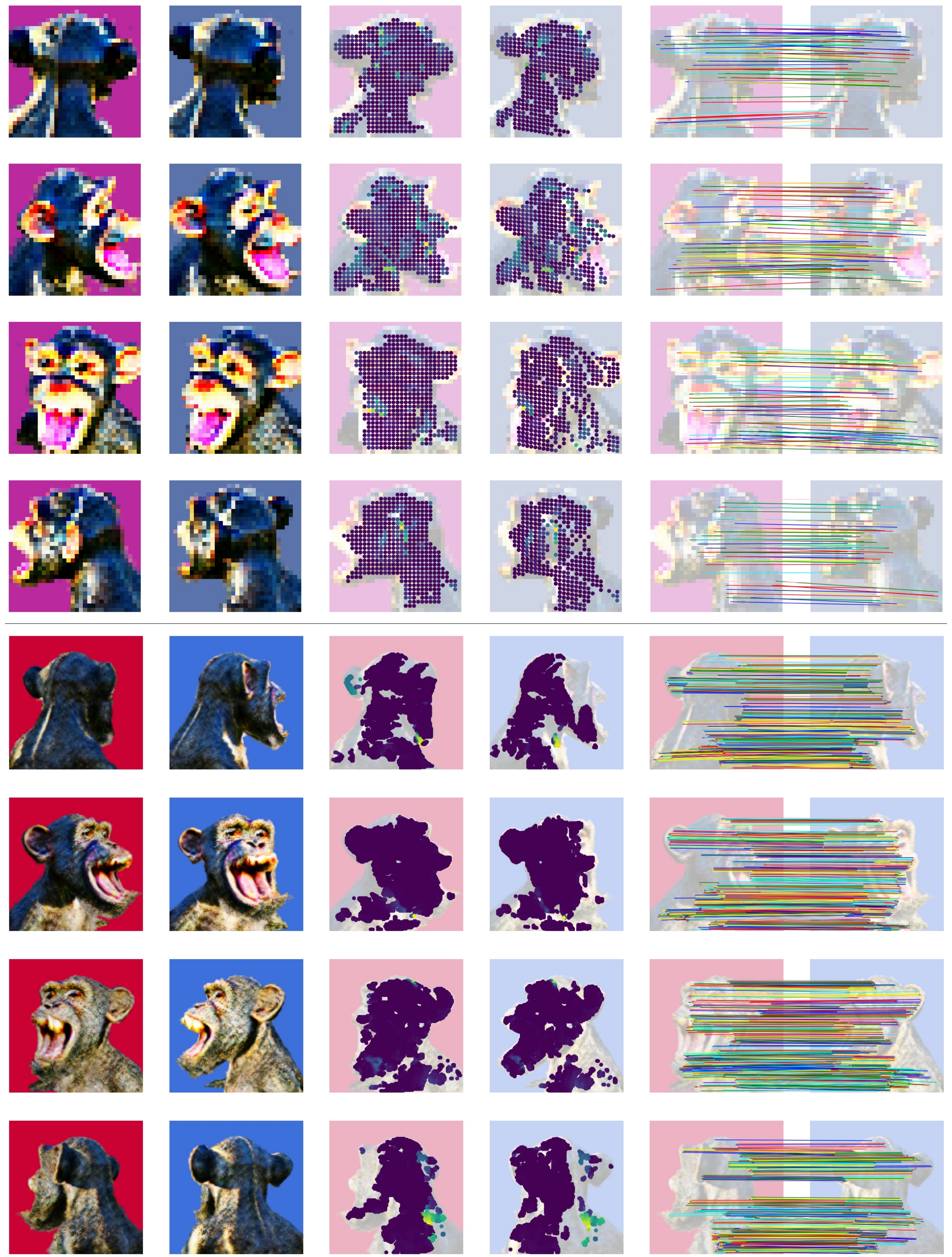}
    \end{center}
    \vspace{-5.0mm}
      \caption{\textbf{Correspondence visualization}. 
      Text prompt - "A chimpanzee with a big grin".
      First two columns show rendered views, and the next two columns visualize the difference between the cross-view correspondences and NeRF reprojections, where brighter colours show higher difference.
      Non-coloured regions show that their correpondences have been filtered out.
      corr$_{\text{Diff}}$ correspondences that have the top 20\% difference from corr$_{\text{NeRF}}$ were visualized on the rightmost two columns.
      The images at top 4 rows were rendered at $32\times32$, and the lower 4 rows were rendered at $128\times128$.
}
\label{fig:supp_corr_viz_1}
\end{figure*}

\begin{figure*}[ht]
    \begin{center}
        \includegraphics[width=0.9\linewidth]{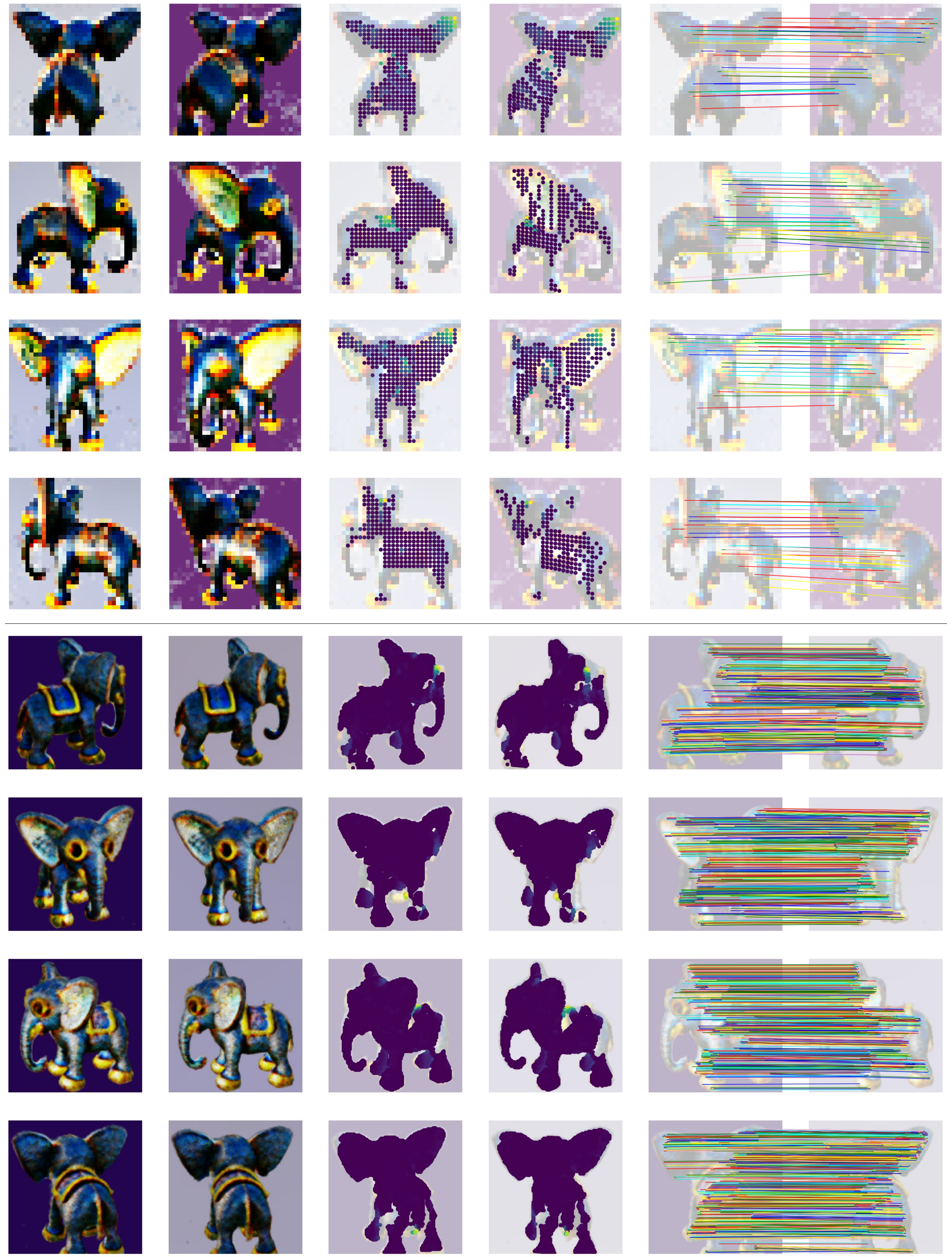}
    \end{center}
    \vspace{-5.0mm}
      \caption{\textbf{Correspondence visualization}. 
      Text prompt - "A cute steampunk elephant".
      First two columns show rendered views, and the next two columns visualize the difference between the cross-view correspondences and NeRF reprojections, where brighter colours show higher difference.
      Non-coloured regions show that their correpondences have been filtered out.
      corr$_{\text{Diff}}$ correspondences that have the top 20\% difference from corr$_{\text{NeRF}}$ were visualized on the rightmost two columns.
      The images at top 4 rows were rendered at $32\times32$, and the lower 4 rows were rendered at $128\times128$.
}
\label{fig:supp_corr_viz_2}
\end{figure*}

\begin{figure*}[ht]
    \begin{center}
        \includegraphics[width=0.9\linewidth]{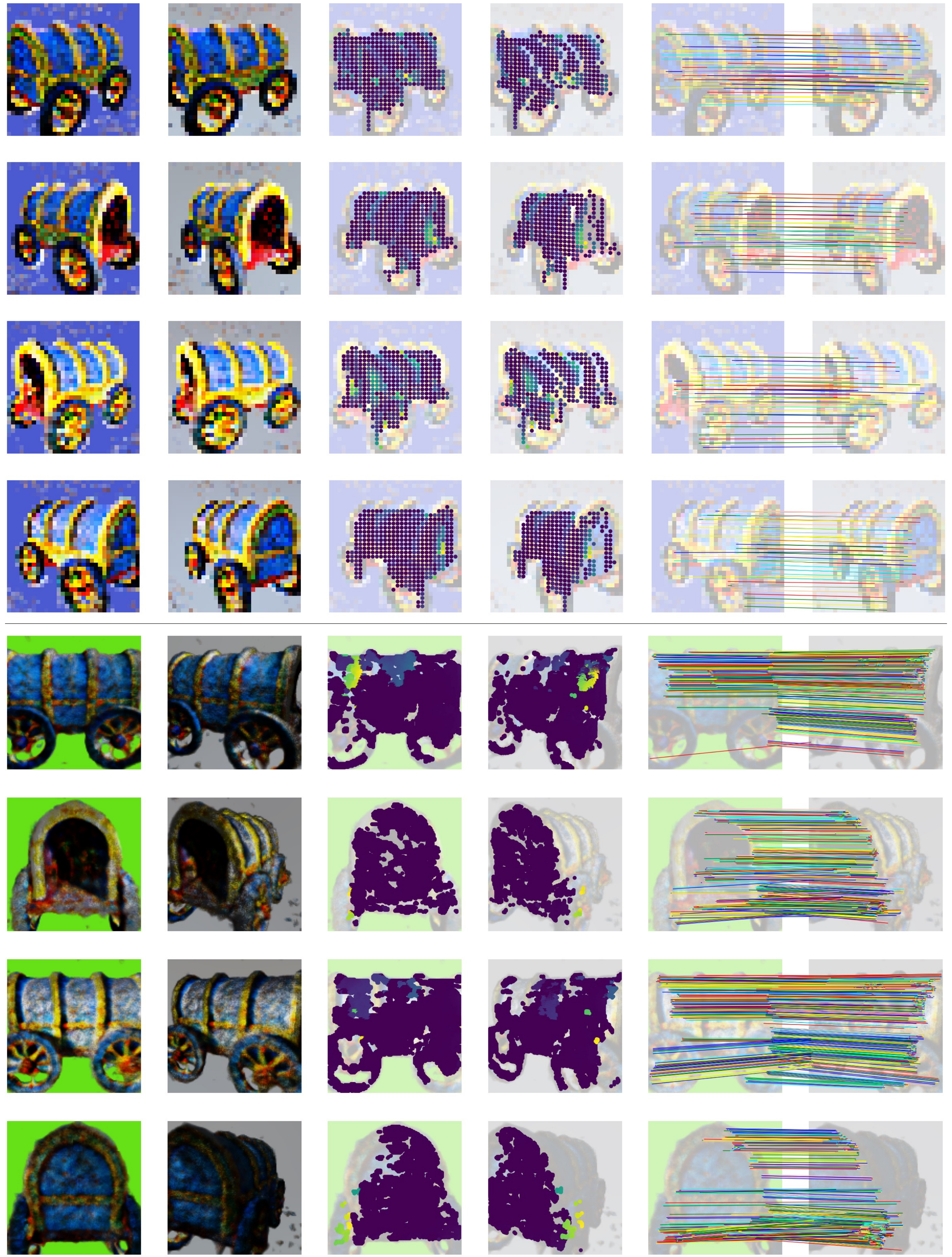}
    \end{center}
    \vspace{-5.0mm}
      \caption{\textbf{Correspondence visualization}. 
      Text prompt - "A DSLR photo of a covered wagon".
      First two columns show rendered views, and the next two columns visualize the difference between the cross-view correspondences and NeRF reprojections, where brighter colours show higher difference.
      Non-coloured regions show that their correpondences have been filtered out.
      corr$_{\text{Diff}}$ correspondences that have the top 20\% difference from corr$_{\text{NeRF}}$ were visualized on the rightmost two columns.
      The images at top 4 rows were rendered at $32\times32$, and the lower 4 rows were rendered at $128\times128$.
}
\label{fig:supp_corr_viz_3}
\end{figure*}

\methodName leverages the annotation-free cross-view correspondences to guide the erroneous NeRF depths, consequently correcting the 3D infidelities.
We provide the visualizations of the correspondences in Figs~\ref{fig:supp_corr_viz_1} to~\ref{fig:supp_corr_viz_3}.
Specifically, the 3rd and 4th columns depicts the disparity between the cross-view correspondences ($\text{corr}_{\text{diff}}$) and the NeRF correspondences from reprojection ($\text{corr}_{\text{NeRF}}$), where brighter regions have higher disparities, \ie, higher chances of infidelities.
The 5th and 6th columns illustrate the cross-view correspondences with the top 20\% disparity values.
While we do not have the ground-truth correspondences to quantitatively evaluate the quality of the cross-view correspondences, it can be visually seen that the cross-view correspondences are coherent to human perception.

\section{Correspondence as geometry pre-/post-processing}
\label{sec:supp_prepostprocessing}
In our current scheme, we are supervising NeRF as our 3D output representation in an \textit{alternating} manner using the cross-view correspondence loss $\mathcal{L}_{\text{corr}}$ and $\mathcal{L}_{\text{SDS}}$, in the \textit{midst} of the NeRF optimization process.

In this section, we provide comparative qualitative results of different schemes of leveraging the cross-view correspondences, (1) using only $\mathcal{L}_{\text{corr}}$ for a fixed number of iterations in the middle of NeRF optimization to fix any errors prior to refining the details of the 3D outputs, and (2) using $\mathcal{L}_{\text{corr}}$ as a post-processing refinement to correct the 3D infidelities after the SDS optimization is completed.

We show the results of this comparative experiment in~\cref{fig:supp_prepostprocessing}, where it can be seen that our current scheme yields the best results in comparison to the pre-processing or post-processing schemes.
For the suboptimal results when using pre-processing, we conjecture this is because the 3D appearance of the output is premature at earlier stages, and using the cross-view correspondence loss at that stage strongly limits the geometric appearance of the output.
This is can be particularly detrimental in the potential presence of any erroneous correspondences, and using the $\mathcal{L}_{\text{corr}}$ alone without $\mathcal{L}_{\text{SDS}}$ may lead to the accumulation of errors.

We also assume this to be the reason behind the failure of using $\mathcal{L}_{\text{corr}}$ solely as a post-processing method.
While the accumulated errors can be somewhat alleviated via the remaining SDS-supervised iterations when using the pre-processing scheme, the post-processing scheme has no way to alleviate the accumulated errors from 2,000 iterations of cross-view correspondences.

\section{Effect of image resolution on output}
\label{sec:supp_image_resolution}
As mentioned in~\cref{sec:experiment}, while MVDream~\cite{shi2023mvdream} finally uses NeRF rendered views at resolutions of 256$\times$256, we use final rendered view resolutions of 128$\times$128.
We observed this maintains the 3D output quality and infidelities, while incurring significantly less latency and memory overhead.
Therefore, we determined that using rendered view resolutions of 128$\times$128 was sufficient to evaluate the efficacy of \methodName while using lower computational resources.

\begin{figure*}[ht]
    \begin{center}
        \includegraphics[width=\linewidth]{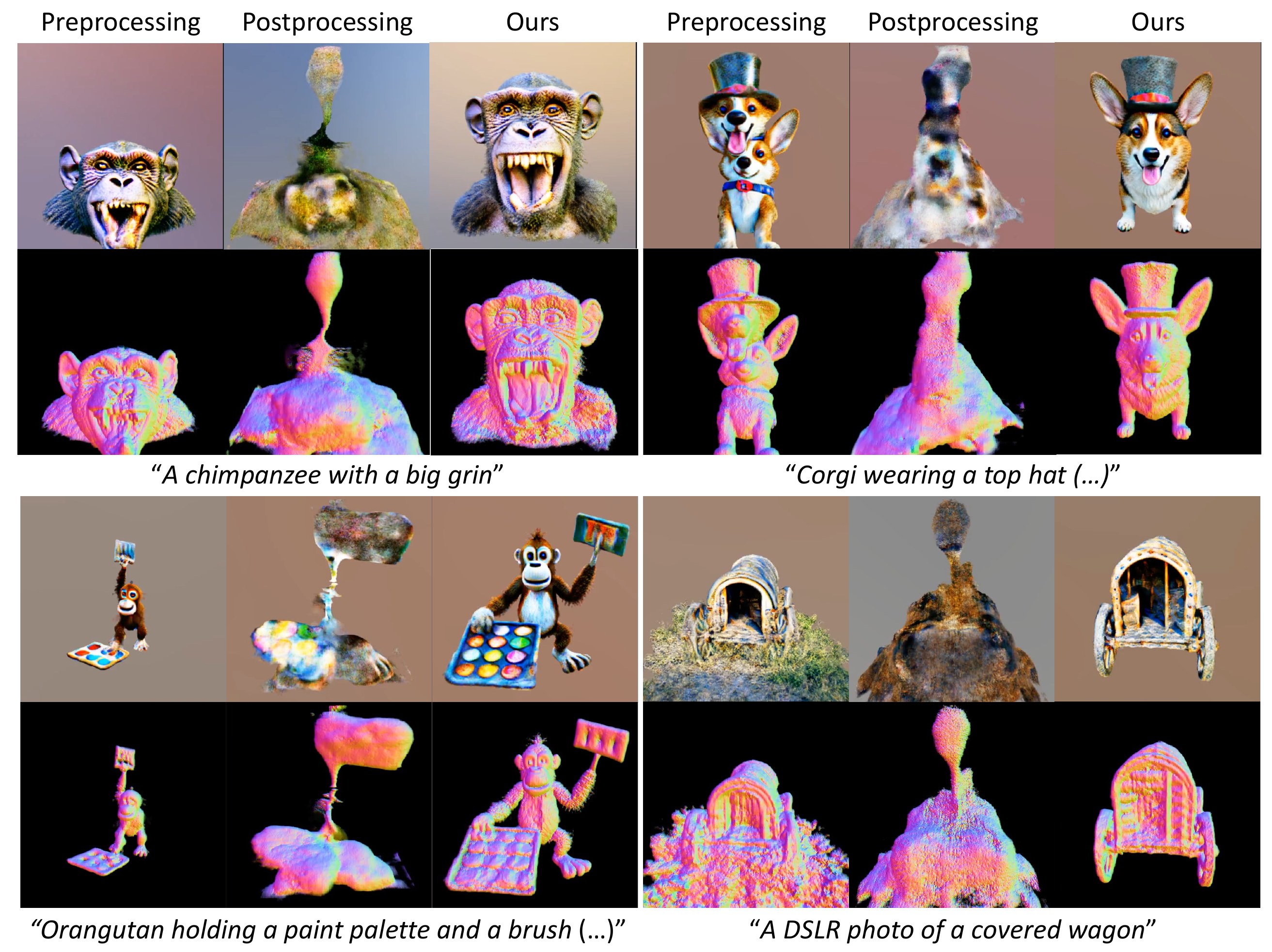}
    \end{center}
    \vspace{-5.0mm}
      \caption{\textbf{Results when using cross-view correspondence loss as a preprocessing / postprocessing step in NeRF optimization}. 
      Instead of using cross-view correspondence loss as a pre-processing scheme (2,000 $\mathcal{L}_\text{corr}$ only iterations after 3,000 iterations, followed by remaining $\mathcal{L}_\textrm{SDS}$ iterations) or post-processing scheme (2,000 $L_\text{corr}$ only iterations after all $\mathcal{L}_\textrm{SDS}$ iterations have finished), our current scheme of alternating supervision yields the best results. 
}
\label{fig:supp_prepostprocessing}

\begin{center}
        \includegraphics[width=\linewidth]{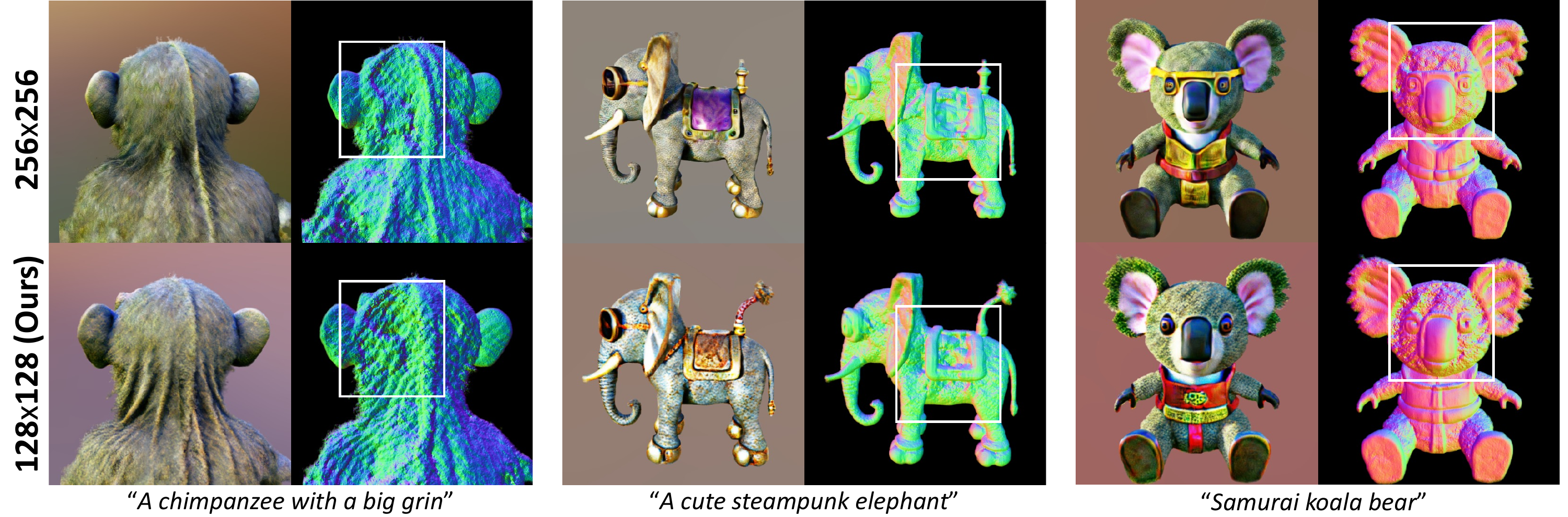}
    \end{center}
    \vspace{-5.0mm}
      \caption{\textbf{Comparison of MVDream~\cite{shi2023mvdream} at different resolutions}. 
      The overall quality of the output, and the 3D infidelities remain even at lower resolutions of 128$\times$128 compared to the original setting of 256$\times$256 of MVDream~\cite{shi2023mvdream}.
      We therefore use resolutions of 128$\times$128 in our experiments to quickly validate the efficacy of \methodName with lower memory and latency overhead.
}
\label{fig:supp_image_resolution}
\end{figure*}

In this section, we qualitatively evidence that even with the lower rendered-view resolutions, the overall output quality and infidelities are nearly consistent in~\cref{fig:supp_image_resolution}.

\section{Progressive visualization}
\label{sec:supp_progressive_visualization}
\begin{figure*}[ht]
    \begin{center}
        \includegraphics[width=0.9\linewidth]{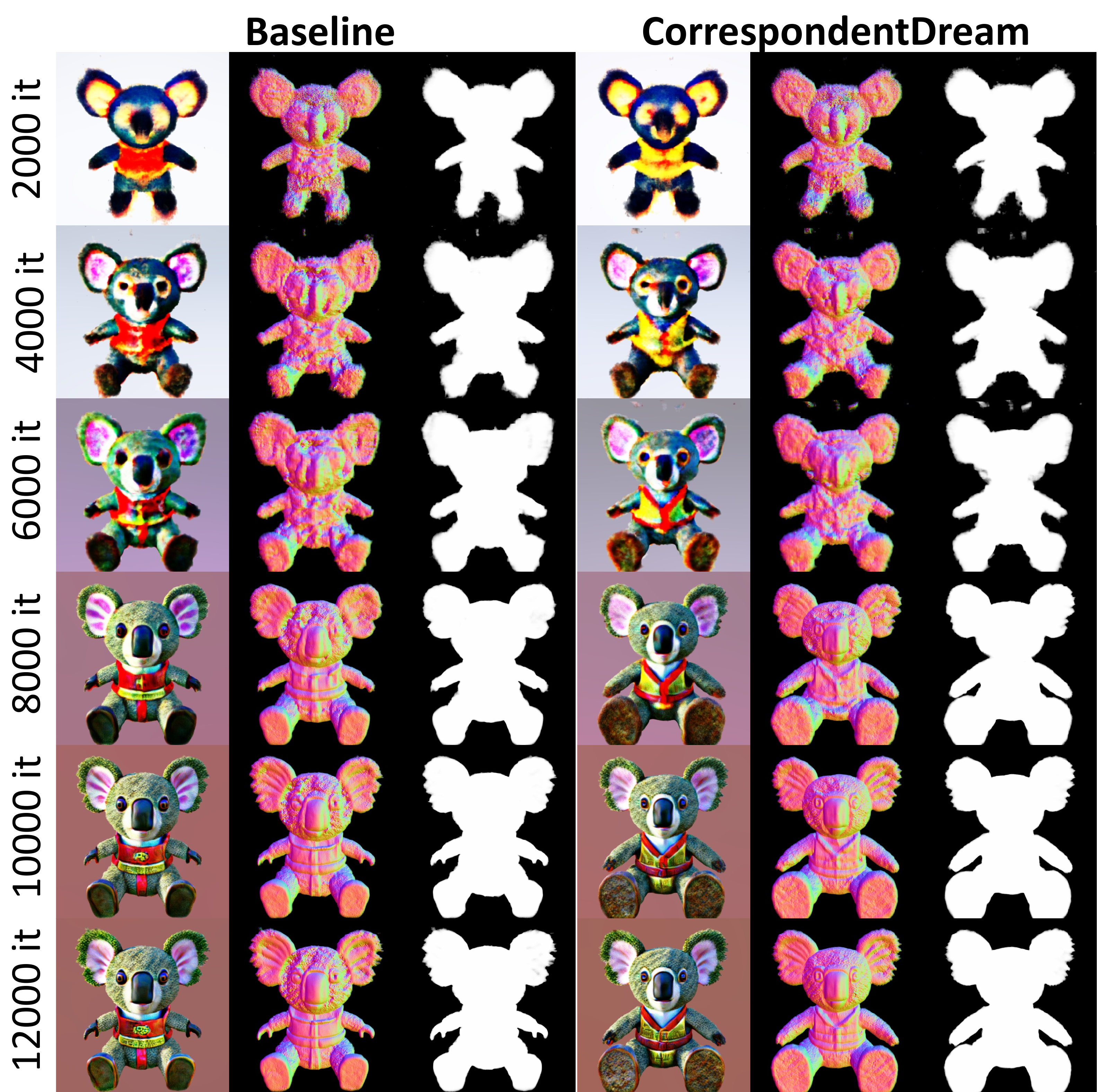}
    \end{center}
    \vspace{-5.0mm}
      \caption{\textbf{Visualization of rendered outputs along NeRF optimization}. 
      We visualize the intermediate and final rendered outputs of the baseline (MVDream~\cite{shi2023mvdream}) and \methodName for a qualitative comparison.
      The text prompt used was "Samurai koala bear".
      It can be seen that the 3D infidelities are corrected along the NeRF optimization of \methodName, whereas the infidelities remain in the baseline.
}
\label{fig:supp_progressive_viz_1}
\end{figure*}

\begin{figure*}[ht]
    \begin{center}
        \includegraphics[width=0.9\linewidth]{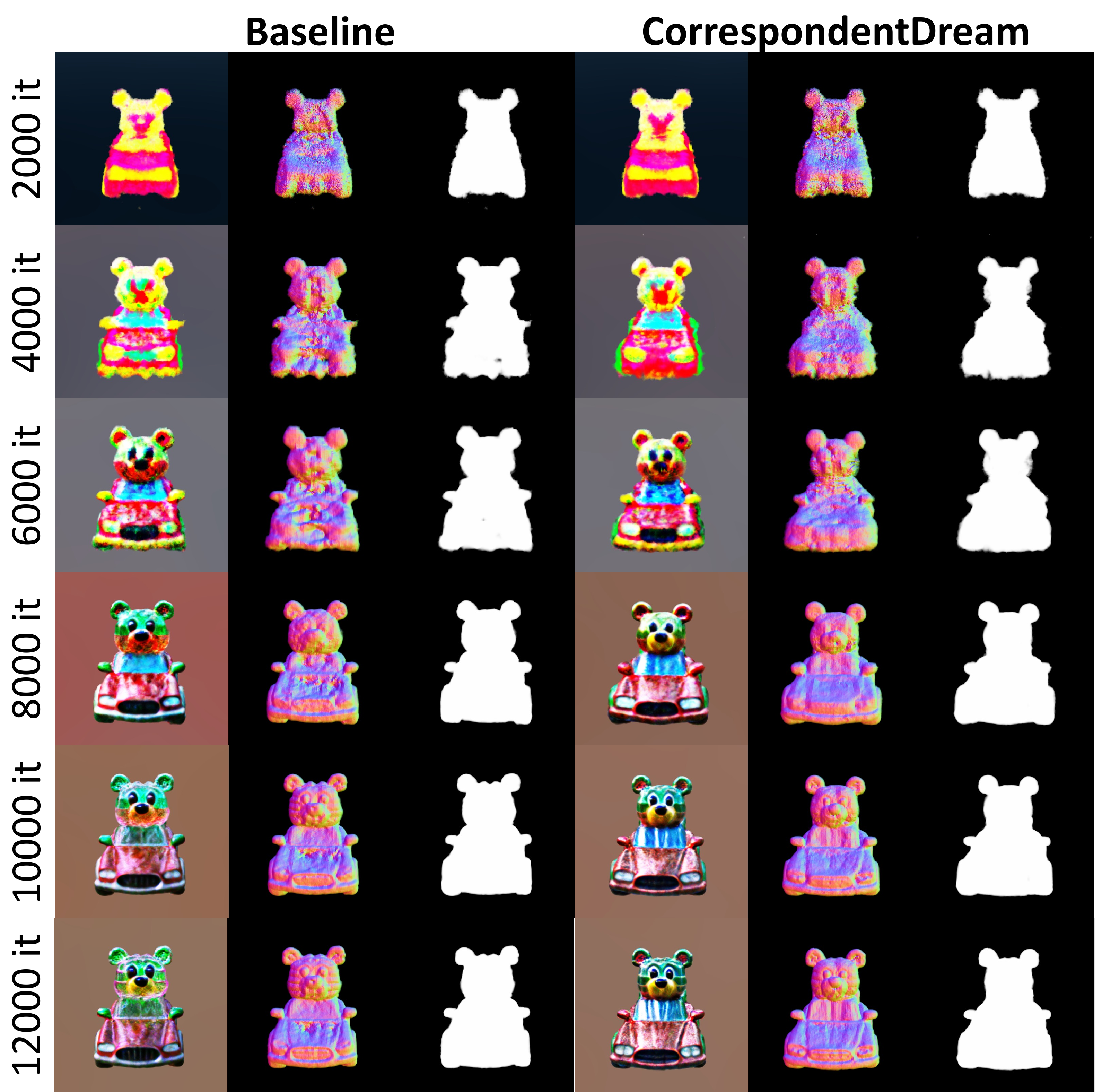}
    \end{center}
    \vspace{-5.0mm}
      \caption{\textbf{Visualization of rendered outputs along NeRF optimization}. 
      We visualize the intermediate and final rendered outputs of the baseline (MVDream~\cite{shi2023mvdream}) and \methodName for a qualitative comparison.
      The text prompt used was "a zoomed out DSLR photo of a gummy bear driving a convertible".
      It can be seen that the 3D infidelities are corrected along the NeRF optimization of \methodName, whereas the infidelities remain in the baseline.
}
\label{fig:supp_progressive_viz_2}
\end{figure*}

\begin{figure*}[ht]
    \begin{center}
        \includegraphics[width=0.9\linewidth]{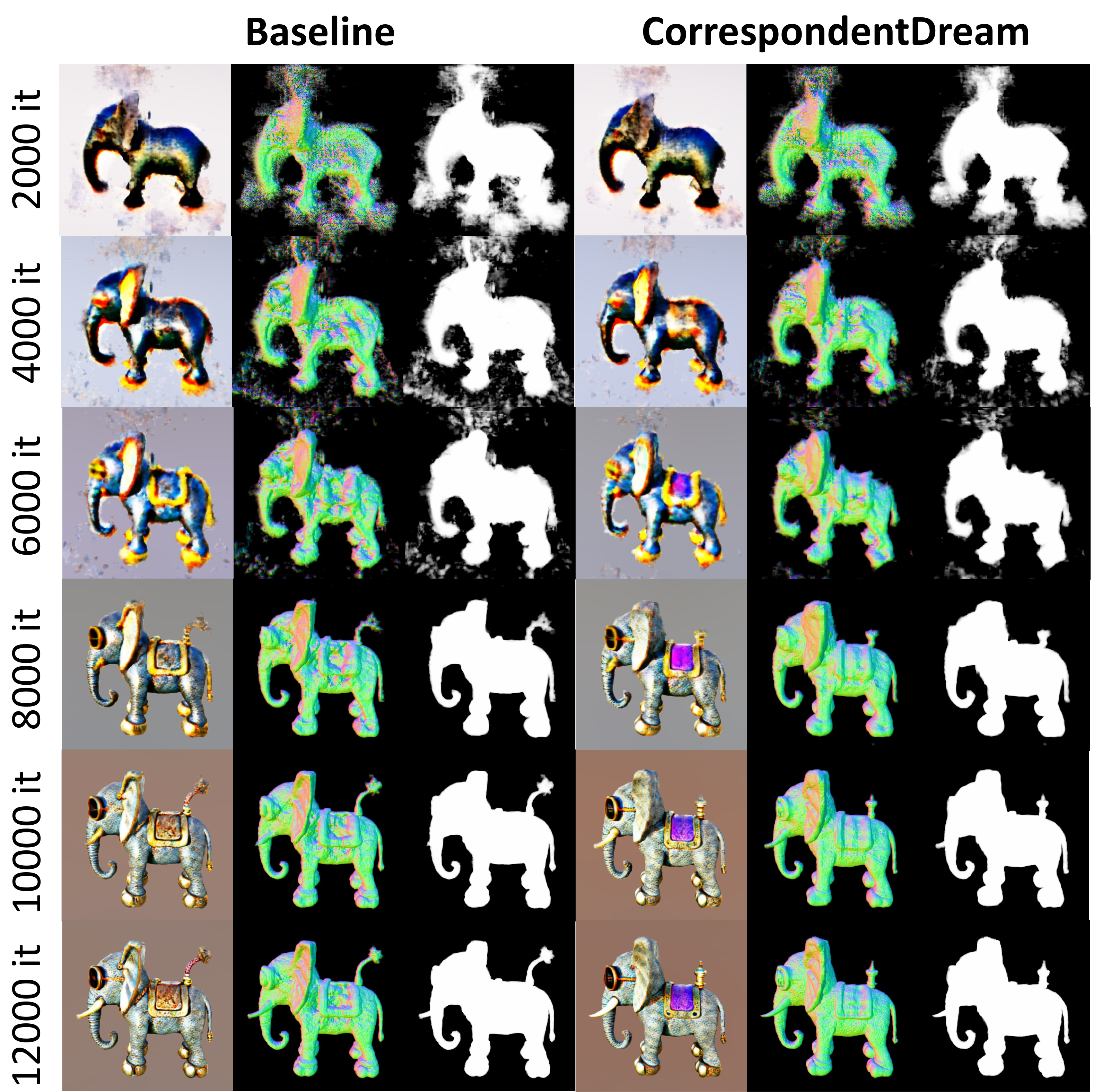}
    \end{center}
    \vspace{-5.0mm}
      \caption{\textbf{Visualization of rendered outputs along NeRF optimization}. 
      We visualize the intermediate and final rendered outputs of the baseline (MVDream~\cite{shi2023mvdream}) and \methodName for a qualitative comparison.
      The text prompt used was "A cute steampunk elephant".
      It can be seen that the 3D infidelities are corrected along the NeRF optimization of \methodName, whereas the infidelities remain in the baseline.
}
\label{fig:supp_progressive_viz_3}
\end{figure*}

\begin{figure*}[ht]
    \begin{center}
        \includegraphics[width=0.9\linewidth]{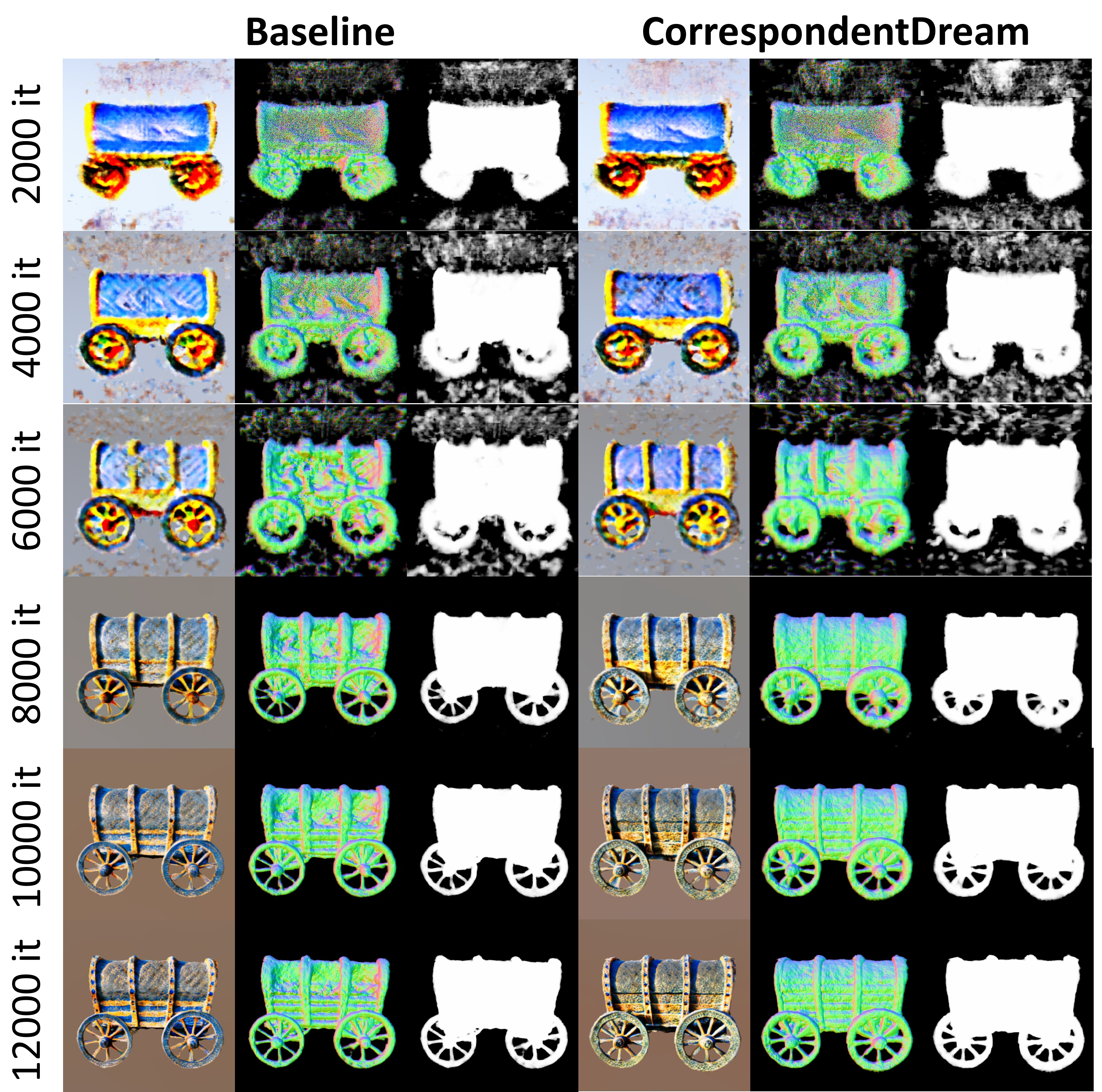}
    \end{center}
    \vspace{-5.0mm}
      \caption{\textbf{Visualization of rendered outputs along NeRF optimization}. 
      We visualize the intermediate and final rendered outputs of the baseline (MVDream~\cite{shi2023mvdream}) and \methodName for a qualitative comparison.
      The text prompt used was "A DSLR photo of a covered wagon".
      It can be seen that the 3D infidelities are corrected along the NeRF optimization of \methodName, whereas the infidelities remain in the baseline.
}
\label{fig:supp_progressive_viz_4}
\end{figure*}

In this section, we provide the progressive visualization of how the NeRF is optimized in \methodName, through 2D rendered views along the course of training.
We provide a comparison with MVDream~\cite{shi2023mvdream} to see how our cross-view correspondence loss shows to correct the 3D infidelities.
The visualizations are shown Figs~\ref{fig:supp_progressive_viz_1} to~\ref{fig:supp_progressive_viz_4}, where it can be seen that \methodName fixes the 3D infidelities along the 3D optimization process, with the help of cross-view correspondences.
This is unlike MVDream~\cite{shi2023mvdream}, where the 3D infidelities remain unresolved.

\section{Applicability to other models}
\label{sec:supp_other_models}

\begin{figure}[h]
    \begin{center}
        \includegraphics[width=\columnwidth]{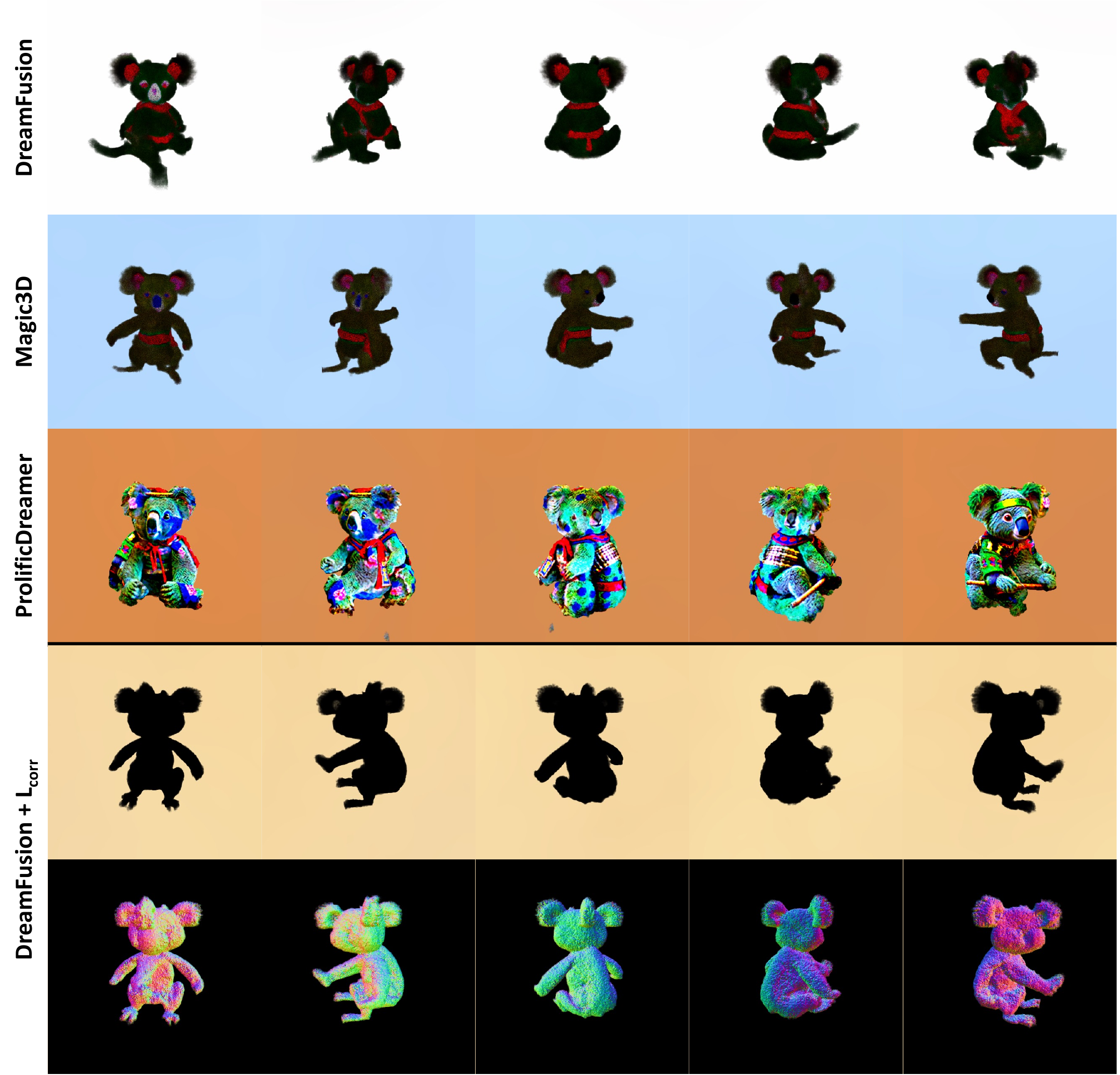}
    \end{center}
      \caption{\textbf{Results of existing text-to-3D generation methods, and applying $\mathcal{L}_\textrm{corr}$ to DreamFusion~\cite{poole2022dreamfusion}.} 
      Existing text-to-3D methods suffer severely from 3D inconsistency \eg Janus face problem, which overwhelms the issue of 3D infidelity.
      The 3D inconsistency makes it challenging to determine accurate cross-view correspondences.
}
\label{fig:rebuttal_applicability}
\end{figure}

\cref{fig:rebuttal_applicability} shows the multi-view rendered results of using the prompt ``Samurai koala bear" for 3 different text-to-3D models (DreamFusion, Magic3D, and ProlificDreamer), where they all suffer from the multi-face problem.
On the last two rows, we also show that applying $\mathcal{L}_{\text{corr}}$ does not alleviate the multi-face problem, and it is thus hard to determine the 3D fidelity of the output.
Thereon, we highlight that while \textbf{\textit{our key problem is the presence of 3D infidelities even when the diffusion prior has good 3D consistency}}, we also \textbf{\textit{rely on good 3D consistency to improve the 3D fidelity}}.
This is because poor 3D consistency causes the 2D renderings to be incorrect, making it challenging to determine accurate cross-view correspondences.
While our method can be integrated with any single/multi-view text/image diffusion priors with \textit{strong 3D consistency}, MVDream was the only such prior at the time of submission.

\section{Computation cost analysis}
\label{sec:computational_cost}

\begin{table}[ht]
\centering
    \begin{tabular}{lcc}
    
    \toprule
    & Memory (GB) & Latency (ms) \\
    \midrule
    8-view low-res $\mathcal{L}_{\text{SDS}}$ & 13.6 & 197\\ 
    2$\times$8-view low-res $\mathcal{L}_{\text{SDS}}$ & 22.5 & 380 \\
    2$\times$8-view low-res $\mathcal{L}_{\text{corr}}$ & 14.1 & 485 \\
    \midrule
    4-view high-res $\mathcal{L}_{\text{SDS}}$ & 14.5 & 198\\ 
    2$\times$4-view high-res $\mathcal{L}_{\text{SDS}}$ & 23.5 & 550 \\
    2$\times$4-view high-res $\mathcal{L}_{\text{corr}}$ & 17.8 & 4700 \\
    \bottomrule
    
    \end{tabular}  
\caption{\textbf{Latency and peak GPU vRAM usage for low-resolution (32$\times$32) and high-resolution (128$\times$128) stages.}
The memory usage of $\mathcal{L}_{\text{corr}}$ is similar to $\mathcal{L}_{\text{SDS}}$ despite rendering twice the number of views.
The latency for $\mathcal{L}_{\text{corr}}$ is higher as we compute a 4D correlation tensor and also perform pre-/post- processing.
}
\label{tbl:rebuttal_computation}
\end{table}

\begin{figure}[h]
    \begin{center}
        \includegraphics[width=\linewidth]{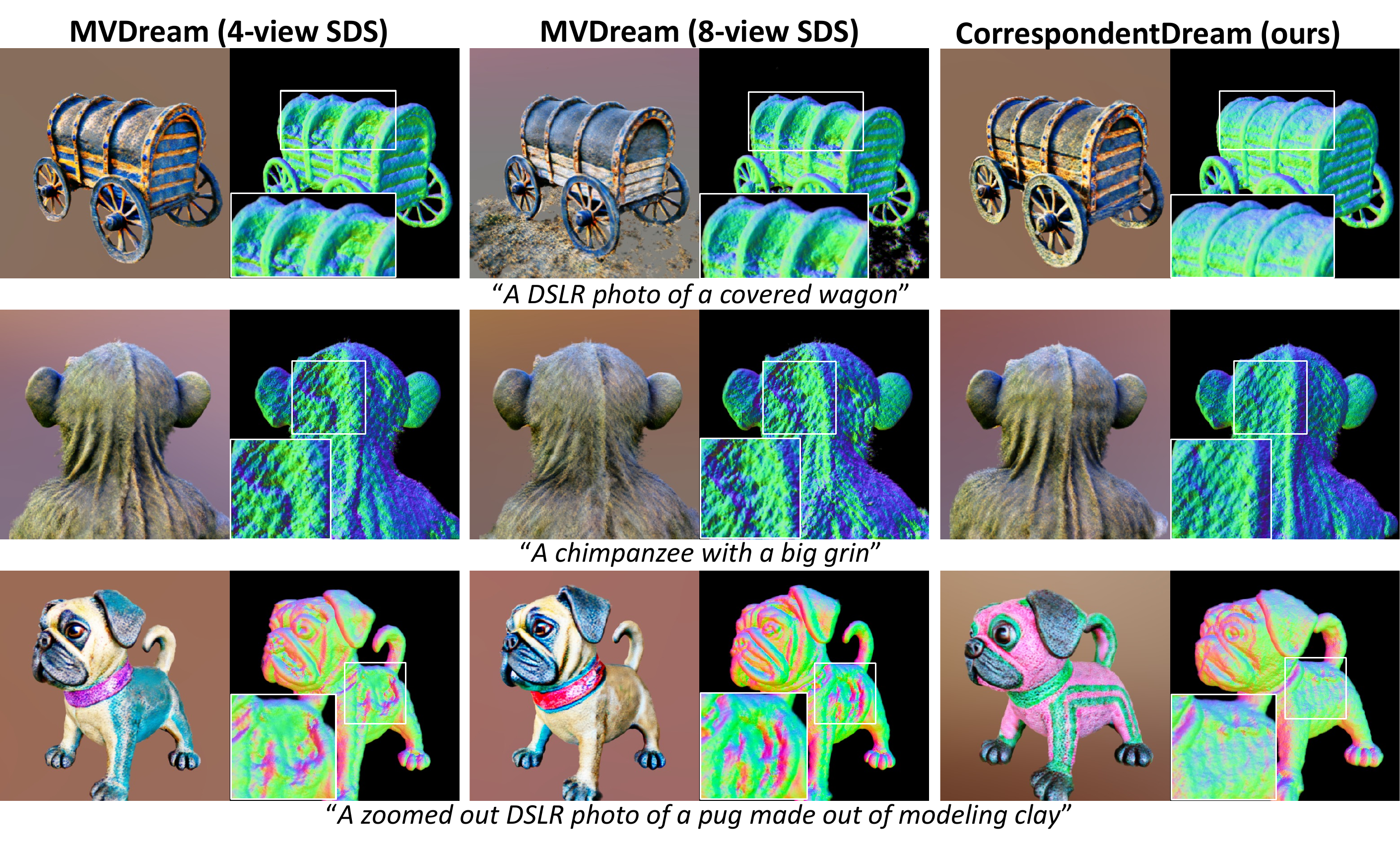}
    \end{center}
\caption{\textbf{Using $\mathcal{L}_{\text{SDS}}$ only but with double the rendered views.} 
Using $\mathcal{L}_{\text{SDS}}$ alone is insufficient to solve the 3D infidelities even with double the number of rendered views as in \methodName.}
\label{fig:rebuttal_8view_sds}
\end{figure}

We show the latency and peak GPU vRAM usage for the low-resolution (32$\times$32) and high-resolution (128$\times$128) stages in~\cref{tbl:rebuttal_computation}.
Compute requirements can differ by text prompt; here we used ``\textit{A zoomed out DSLR photo of a pug made out of modeling clay}".
Specifically, the memory usage of $\mathcal{L}_{\text{corr}}$ is similar to $\mathcal{L}_{\text{SDS}}$ despite having to render twice the number of views.
However, the latency for $\mathcal{L}_{\text{corr}}$ is higher, as we compute a 4D correlation tensor, and also perform pre-/post- processing to obtain reliable correspondences.
We show in~\cref{fig:rebuttal_8view_sds} that $\mathcal{L}_{\text{SDS}}$ with increased number of views is still insufficient to alleviate the 3D infidelities, evidencing the efficacy of our method.

\begin{figure}[h] 
    \centering
        \includegraphics[width=\columnwidth]{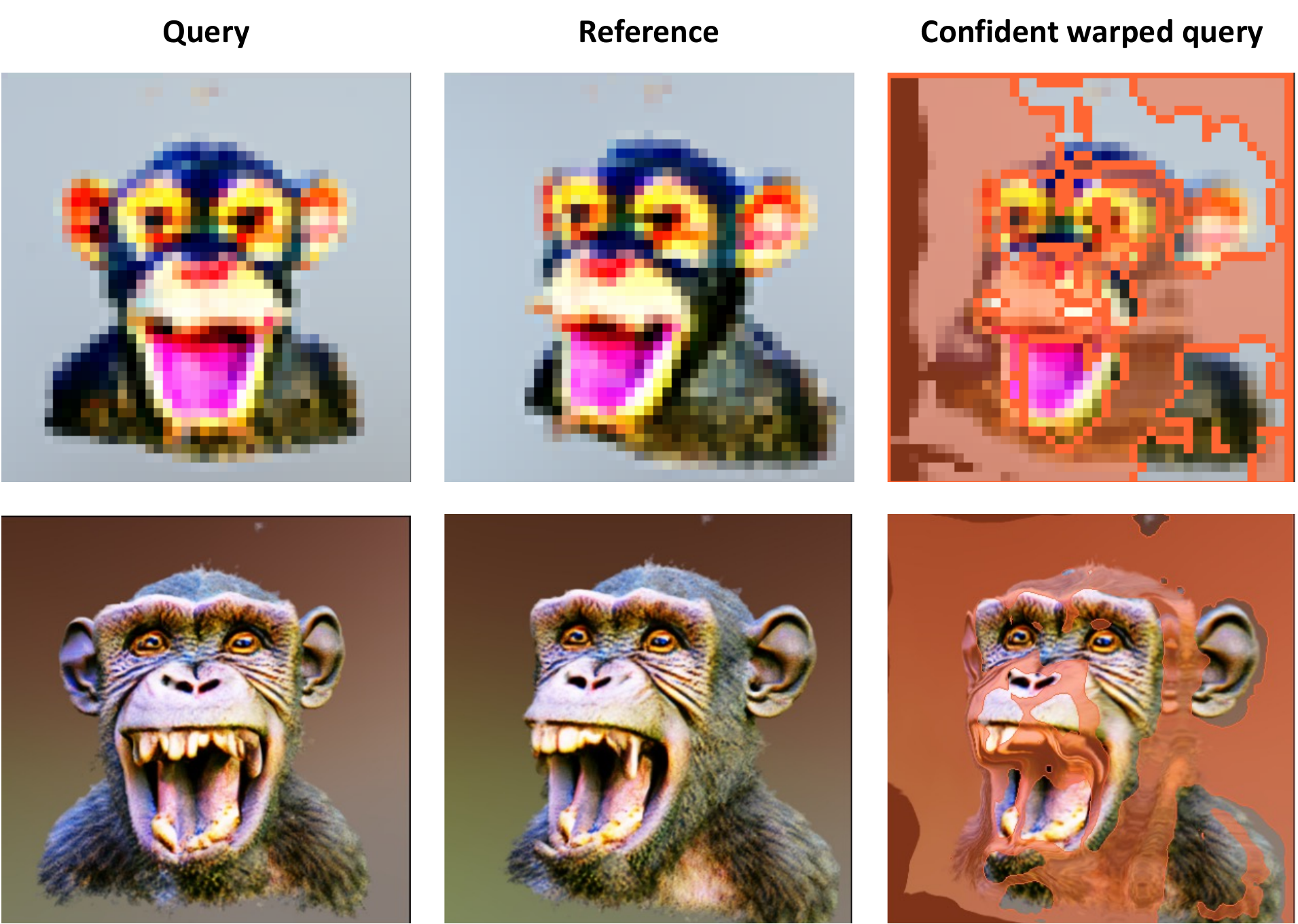}
\caption{\textbf{Visualization of correspondences computed using PDCNet~\cite{truong2021learning}.}
PDCNet fails to find high-confidence correspondences for most of the foreground regions.
}
\label{fig:rebuttal_pdcnet}
\end{figure}

\section{Why not use off-the-shelf matchers?}
\label{sec:rebuttal_offtheshelf_matcher}
The capabilities of off-the-shelf matchers rely heavily on the dataset they were trained on, which is problematic where the domain of  generated 3D object depends on the text prompt.
We visualize the results of warping our i) low-resolution intermediate renderings and ii) high-resolution final rendering using PDCNet~\cite{truong2021learning}\footnote{We used the official code and MegaDepth-pretrained weights.} predictions in~\cref{sec:rebuttal_offtheshelf_matcher}.
Note that PDCNet was the off-the-shelf image matcher which was used in SPARF~\cite{truong2023sparf}.
It can be seen that PDCNet fails to find high-confidence correspondences for most of the foreground regions (shown in orange), especially in the low-resolution renderings.
Also, off-the-shelf methods incur additional computation; PDCNet incurs approximately 1GB memory usage and 4000ms latency to establish correspondences between an image pair.
Using diffusion features eliminates the domain issue without explicit priors or additional compute.
However, we believe that a carefully trained matcher could be more effective at handling diffusion features' shortcomings.

\section{Example prompts}
\label{sec:supp_example_prompts}

In this section, we provide some of the prompts which were used to generate the qualitative examples in the main paper and this supplementary material, and the other prompts which were used in our experiments as well in~\cref{tab:example_prompts}.
The prompts were largely borrowed from DreamFusion~\cite{poole2022dreamfusion} and MVDream~\cite{shi2023mvdream}.

\begin{table*}[h]
\small
\footnotesize
\begin{tabular}{l}
\hline
a bichon frise wearing academic regalia \\
a capybara wearing a top hat, low poly \\
a cat with a mullet \\
a ceramic lion \\
a chimpanzee with a big grin \\
a cute steampunk elephant \\
a DSLR photo of a bear dressed in medieval armor \\
a DSLR photo of a beautiful violin sitting flat on a table \\
a DSLR photo of a corgi lying on its back with its tongue rolling out \\
a DSLR photo of a covered wagon \\
a DSLR photo of a mug of hot chocolate with whipped cream and marshmallows \\
a DSLR photo of an iguana holding a balloon \\
a DSLR photo of a pomeranian dog \\
a DSLR photo of a porcelain dragon \\
a DSLR photo of a puffin standing on a rock \\
a DSLR photo of a pug made out of metal \\
a DSLR photo of a turtle standing on its hind legs, wearing a top hat and holding a cane \\
a DSLR photo of a very cool and trendy pair of sneakers, studio lighting \\
a DSLR photo of a vintage record player \\
a DSLR photo of cat wearing virtual reality headset in renaissance oil painting high detail caravaggio \\
An anthropomorphic tomato eating another tomato \\
an orangutan holding a paint palette in one hand and a paintbrush in the other \\
a wide angle DSLR photo of a colorful rooster \\
a yellow schoolbus \\
a zoomed out DSLR photo of a baby dragon \\
a zoomed out DSLR photo of a colorful camping tent in a patch of grass \\
a zoomed out DSLR photo of a corgi wearing a top hat \\
a zoomed out DSLR photo of a dachsund wearing a boater hat \\
a zoomed out DSLR photo of a gummy bear driving a convertible \\
a zoomed out DSLR photo of a hippo made out of chocolate \\
a zoomed out DSLR photo of an origami bulldozer sitting on the ground \\
a zoomed out DSLR photo of a pug made out of modeling clay \\
a zoomed out DSLR photo of a wizard raccoon casting a spell \\
a zoomed out DSLR photo of a yorkie dog dressed as a maid \\
an astronaut riding a horse \\
Samurai koala bear \\
a DSLR photo of an eggshell broken in two with an adorable chick standing next to it \\
Darth Vader helmet, highly detailed \\
Pikachu with hat \\
A product photo of a toy tank \\
a boy in mohawk hairstyle, head only, 4K, HD, raw \\
Wall-E, cute, render, super detailed, best quality, 4K, HD \\
slayer, assassin with sword, portrait, game, unreal, 4K, HD \\
an alien monster that looks like an octopus, game, character, highly detailed, photorealistic, 4K, HD \\
mushroom boss, cute, arms and legs, big eyes, game, character, render, best quality, super detailed, 4K, HD \\
pentacle sign, 4k, HD\\
\hline
\end{tabular}
\caption{\textbf{Example prompts.} These prompts were largely borrowed from DreamFusion~\cite{poole2022dreamfusion} and MVDream~\cite{shi2023mvdream}.
}
\label{tab:example_prompts}
\end{table*}

\clearpage

{
    \small
    \bibliographystyle{ieeenat_fullname}
    \bibliography{01_main}
}

\end{document}